\documentclass[a4paper]{cas-sc}

\usepackage[numbers,sort&compress]{natbib}

\usepackage{calc}
\usepackage{tabularx}
\usepackage{mathtools}
\usepackage{bm}
\usepackage{microtype}
\usepackage{url}

\newif\ifarxiv
\arxivtrue

\ExplSyntaxOn
\keys_define:nn { cas / fig }
  { unknown .code:n = { \tl_set:Nx \l_fig_pos_tl { \l_keys_key_str } } }
\keys_define:nn { cas / tbl }
  { unknown .code:n = { \tl_set:Nx \l_tbl_pos_tl { \l_keys_key_str } } }
\ExplSyntaxOff

\graphicspath{{../../../../local-image-archive/journals/2026/neural-dynamics-model/}{../../../ImageArchive/journals/2026/neural-dynamics-model/}{../els-cas-templates/}{./image/}}

\newcolumntype{L}[1]{>{\raggedright\arraybackslash}p{#1}}
\newcolumntype{C}[1]{>{\centering\arraybackslash}p{#1}}
\newcolumntype{Y}{>{\raggedright\arraybackslash}X}

\newcommand{\R}{\mathbb{R}}

\begin{document}
\let\WriteBookmarks\relax
\def\floatpagepagefraction{1}
\def\textpagefraction{.001}

\shorttitle{Neural Reduced Dynamics for Complex Robot Control}

\shortauthors{H. Zhang and D. Negrut}

\title[mode = title]{Learning the Right Abstraction: Neural Reduced Dynamics for Complex Robot Control}

\tnotemark[1]
\tnotetext[1]{This work was supported in part by the Simulation-Based Engineering Lab (SBEL) at the University of Wisconsin--Madison.}

\author[1]{Harry Zhang}
\cormark[1]
\ead{hzhang699@wisc.edu}
\credit{Conceptualization, Methodology, Software, Investigation, Writing -- original draft}

\affiliation[1]{organization={Department of Mechanical Engineering, University of Wisconsin--Madison},
            city={Madison},
            postcode={53706},
            state={WI},
            country={USA}}

\author[1]{Dan Negrut}
\ead{negrut@wisc.edu}
\credit{Conceptualization, Supervision, Funding acquisition, Writing -- review \& editing}

\cortext[1]{Corresponding author}

\begin{abstract}
High-fidelity embodied AI simulators provide realistic evaluation of complex robotic systems, but their computational cost limits their direct use for large-scale reinforcement learning campaigns. We advocate the use of less accurate but more expeditious simulations, which might draw on data-driven, e.g., neural dynamics, models. This contribution argues that the practical value of a neural dynamics model for complex robot control lies in learning the \emph{right abstraction}: a reduced state that preserves the control-relevant physics of the high-fidelity system while enabling high-throughput policy learning.
We develop a neural reduced dynamics (NRD) framework that separates the state the model propagates from what can be supplied as an input or recovered analytically, trains policies entirely inside the frozen learned model, and validates them back in the high-fidelity simulator.
Two case studies instantiate it across three control tasks: terrain-aware HMMWV trajectory tracking on rigid, bumpy and deformable Continuum Representation Model (CRM) terrain; and goal reaching for a stock tracked vehicle and its front-mounted articulated arm. Every policy transfers back to the high-fidelity simulator. A single policy trained inside the terrain-conditioned dynamics model, and given no terrain input of its own, attains lower median and mean tracking error than both single-terrain specialists on all three terrains, including zero-shot bumpy terrain. Quantitatively, the tracked vehicle reaches $100$ of $100$ goals and the arm $97$ of $100$, with zero contacts or joint-limit violations. The NRD models advance roughly four orders of magnitude faster in simulated time than the high-fidelity simulator scenes they replace, making iterative on-policy learning practical and supporting neural reduced dynamics as a bridge between accurate but expensive physics simulation and scalable robot learning.
\end{abstract}



\maketitle

\begin{figure*}[t]
  \centering
  \includegraphics[width=\textwidth]{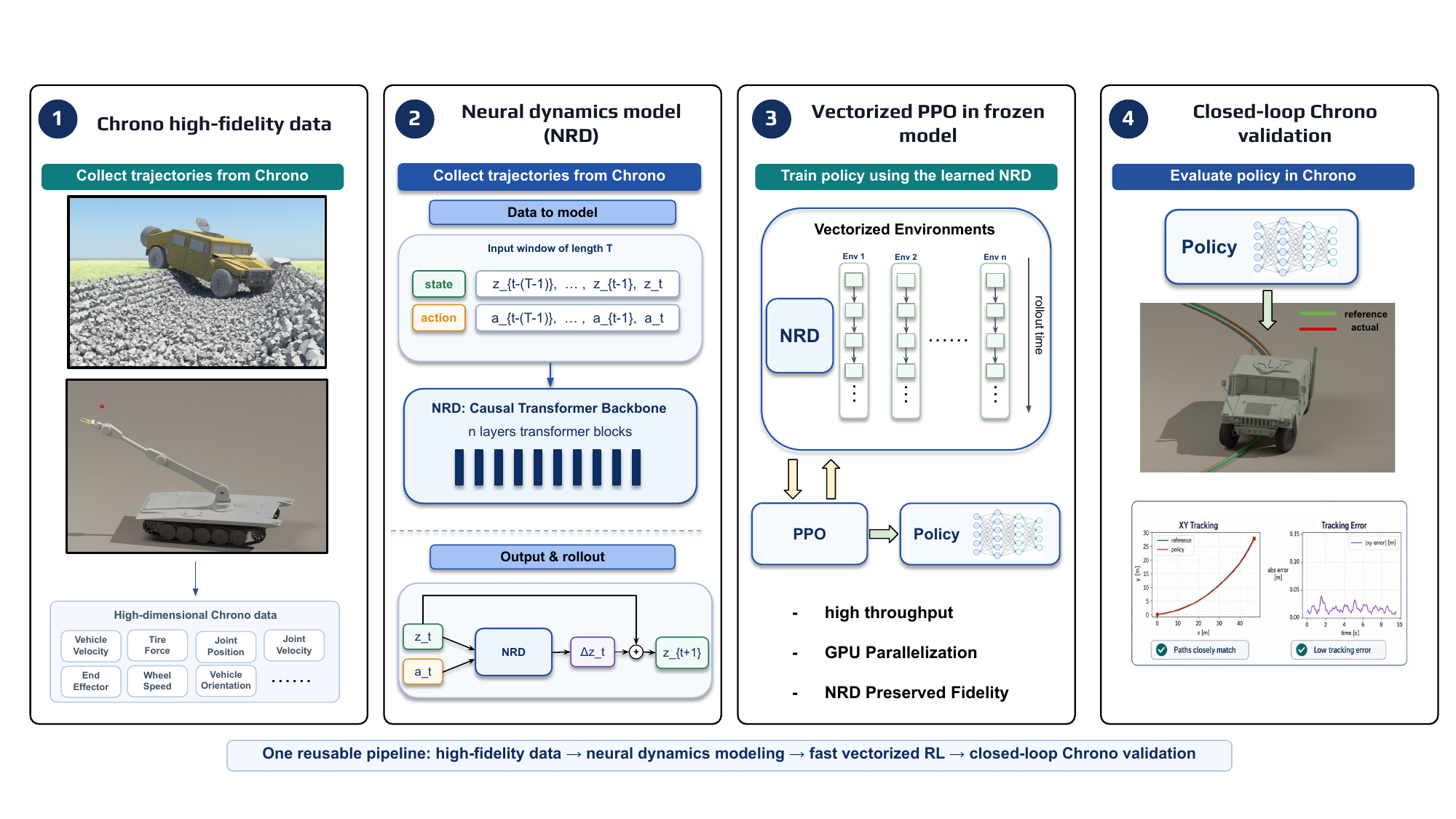}
  \caption{Overview of the neural reduced dynamics framework. A single reusable
  pipeline turns expensive high-fidelity physics into scalable policy learning:
  (1)~high-fidelity Chrono simulation generates state--action trajectories for
  complex robots (a terrain-traversing HMMWV and a tracked vehicle with an
  articulated arm), yielding high-dimensional signals such as vehicle velocity,
  tire force, joint position/velocity, end-effector position, wheel speed, and
  vehicle orientation; (2)~a causal-transformer neural reduced dynamics (NRD) model
  learns next-step reduced dynamics from a window of state--action
  history; (3)~vectorized PPO trains a control policy entirely inside the frozen,
  high-throughput NRD model; and (4)~the resulting policy is validated closed-loop
  back in high-fidelity Chrono, where it is evaluated closed-loop on the original
  control task.}
  \label{fig:overview}
\end{figure*}

\section{Introduction}
\label{sec:introduction}

Complex robots, e.g., wheeled vehicles crossing deformable terrain, tracked
platforms, and articulated manipulators, are governed by multibody dynamics,
frictional and deformable contact, and actuator behavior that resist compact
analytical description. High-fidelity embodied AI simulators such as Project
Chrono \cite{chronoOverview2013,chronoOverview2016} resolve these effects and provide a controlled, repeatable environment in
which a candidate controller can be evaluated before hardware
trials~\cite{chronoVehicle2019}. That fidelity is expensive. A detailed granular-terramechanics or
tracked-vehicle scene can run several times slower than real time, while reinforcement learning (RL) consumes millions of environment transitions across repeated training runs. Training a policy directly inside such a simulator is impractical for the scenarios where high fidelity matters most. High-fidelity simulation supplies the physics needed to evaluate a controller, but not the throughput needed to search for one efficiently.

A learned dynamics model is one common remedy for this cost, see, for example, \cite{justin-FNODE-2026}, but introducing
one does not by itself resolve the difficulty; the open question is what the
model should represent. Learning the transition of the complete simulator state is neither
necessary nor easy: many internal degrees of freedom need not be represented
explicitly for the control task at hand, some quantities are deterministic
functions of others, and contact and deformable-terrain states can be extremely
high-dimensional. An aggressively simplified surrogate has the opposite failure
mode. It runs quickly, but may discard the
tire slip, load transfer, or actuator lag on which the controller depends, so
that a policy trained in it may fail when transferred to the high-fidelity
dynamics.
Neither low one-step prediction error nor generic state compression establishes
that a surrogate is fit for policy learning. This raises the central question of
the paper: what should a neural reduced dynamics (NRD) model retain, receive as input,
reconstruct analytically, and omit, so that it is both fast and useful for
control?

We argue that a useful NRD model should preserve
task-relevant fidelity rather than minimize state dimension for its own sake.
It should propagate the recurrent variables on which the task-relevant dynamics
depend, treat externally specified commands and regime information as inputs
where appropriate, and recover quantities that follow from the propagated state
through known kinematic relationships outside the learned transition. Figure~\ref{fig:overview} shows the
resulting pipeline, which is shared by both case studies. Excited trajectories are
collected from Chrono; a history-based NRD model built
on a causal transformer~\cite{vaswani2017attention} is trained on a task-specific
reduced state and selected by open-loop rollout fidelity rather than one-step
loss; the selected model is frozen and vectorized so that a policy can be trained
against thousands of copies with proximal policy
optimization (PPO)~\cite{schulman2017proximal}; and the trained policy is returned to
the full Chrono system for closed-loop evaluation. Here \emph{reduced} denotes
preserved task-relevant fidelity, not minimal dimensionality, and closed-loop
transfer to Chrono is the decisive criterion, while the NRD model's supervised error
and the policy's reward inside it serve as intermediate evidence.

We instantiate this framework in two complementary case studies. Case Study~I
builds a terrain-conditioned NRD model for the HMMWV. Its reduced state keeps
body dynamics, tire normal loads, and wheel speeds, and an explicit terrain
context tells the model whether the ground is rigid or deformable. A single
generalist policy is trained inside that model. Back in Chrono, that policy
attains the lowest median and mean closed-loop tracking error among the policies
compared here, on rigid flat terrain, on CRM deformable soil \cite{weiGranularSPH2021,Huzaifa2026CRM}, and on bumpy rigid terrain it
never saw during training. Each single-terrain specialist, in contrast, degrades
sharply off its training terrain. Case Study~II applies the same pipeline to two substantially
different abstractions on one tracked-vehicle platform: a three-dimensional planar state
for tracked-vehicle goal reaching and an eight-dimensional joint-space state for arm
reaching, with vehicle pose integrated and end-effector position recovered through
forward kinematics outside the NRD model. Policies trained entirely inside these frozen NRD models transfer
to Chrono, reaching $100$ of $100$ vehicle goals within $0.75$\,m and $97$ of $100$
arm goals within $0.05$\,m, with zero contacts or joint-limit violations across
the arm rollouts. Across both cases, the batched NRD models achieve measured
aggregate simulated-time throughput ratios of up to $58{,}400\times$ under the
reported GPU-batched surrogate and single-process Chrono configurations, making
iterative on-policy RL practical.

Our contribution is methodological. The paper studies
how control-relevant state, applied actions, regime context, and known analytical
structure should be organized within a learned surrogate, and evaluates that
organization through downstream policy transfer. The causal transformer and PPO
implementation provide just one realization of the framework rather than the central
contribution. Specifically, the paper makes the following contributions.

\begin{itemize}
  \item \textbf{Task-specific neural reduced-dynamics framework.} A reusable
  ``Chrono-data''--to--``policy'' pipeline that separates recurrent state, applied actions,
  optional context, and analytically derived quantities; selects models by
  held-out rollout fidelity; freezes and vectorizes them for policy training; and
  validates the resulting policies through closed-loop Chrono transfer.
  \item \textbf{Terrain-conditioned HMMWV dynamics and control.} A
  fifteen-dimensional terramechanics abstraction with a rigid/CRM context code,
  supported by reduced-state and context ablations, whose generalist policy
  attains the lowest median and mean tracking error among the compared policies on
  flat, CRM deformable, and zero-shot bumpy terrain.
  \item \textbf{Cross-task applicability of task-specific abstractions.} Distinct
  three-dimensional tracked-vehicle and eight-dimensional arm abstractions on one
  vehicle-plus-arm platform, producing policies that reach $100/100$ and $97/100$
  Chrono goals, respectively.
\end{itemize}
\section{Background and Related Work}
\label{sec:related-work}

\subsection{Learned Dynamics and Neural Surrogates for Control}
\label{subsec:rw-learned-dynamics}

Learned transition models build off predicted experience rather than costly interaction
with a physical system or a high-fidelity simulator. Probabilistic ensembles with
trajectory sampling (PETS) pair learned dynamics with sampling-based
model-predictive control, propagating model uncertainty into planning to obtain
sample-efficient control~\cite{chua2018pets}. The accompanying risk is
autoregressive model bias: as a model is advanced on its own predictions, error
compounds and an optimizer begins to exploit the model rather than the system it
stands for. Model-based policy optimization (MBPO) manages this by branching short
synthetic rollouts from real states, bounding how far the model is
trusted~\cite{janner2019mbpo}. The learned model here occupies a more demanding
role: it is the entire environment in which the policy is trained, advanced over
complete episodes, so its behavior under sustained recursion is the property that
matters. That role dictates the evidence reported below: one-step accuracy,
autoregressive open-loop rollout against recorded Chrono trajectories, and
closed-loop policy transfer back to Chrono.

Latent world models address the same cost problem from a different starting point,
learning compact dynamics directly from high-dimensional observations and
optimizing behavior inside the learned model. World Models train a controller
inside an imagined environment~\cite{ha2018worldmodels}, PlaNet learns a latent
state-space model from images and plans through imagined
trajectories~\cite{hafner2019planet}, and Dreamer learns actor and value functions
from latent imagination rather than planning in the original
environment~\cite{hafner2020dreamer}; recent work pretrains such a model in
simulation and adapts only its dynamics component with real
observations~\cite{levy2026simdist}. Common to this category is that the
representation itself must be discovered: raw observations do not name physical
quantities, so an encoder is learned jointly with the dynamics. For this work, we
start from the opposite situation. The simulator already exposes variables with
physical identity---body velocities, tire normal forces, wheel speeds, joint
angles---so the central representation problem is physical-variable selection and
validation rather than perceptual encoding:
which variables must be propagated recurrently, which can be supplied as inputs,
and which follow from known kinematics. We freeze the selected model rather than
adapt it, and the transfer we evaluate is sim-to-sim.

The closest contemporary comparisons are neural surrogates built for robot
dynamics. Neural Robot Dynamics learns robot-centric dynamics with learned
contact-solver backends and uses the result as a neural simulation engine for
policy training~\cite{xu2025nerd}. Neighboring efforts are shaped by their
downstream use: smooth neural surrogates regularize sensitivity so that contact
derivatives remain usable for gradient-based legged MPC~\cite{moore2026snsmpc};
on-the-fly adaptation updates part of an incremental dynamics model online under
changing conditions~\cite{altawaitan2026adapting}; and contact-aware neural
dynamics predicts upcoming contacts and refines them with real tactile
data~\cite{jing2026contactaware}. The framework developed here learns an offline,
task-specific reduced transition rather than a general contact solver, and freezes
and vectorizes it for model-free policy optimization rather than differentiating
through it or adapting it online. What separates these approaches is which state is retained, how the model is used
during rollout, and what evidence is taken to show that it suffices for that use.

\subsection{Reduced-Order and Structured Dynamics Models}
\label{subsec:rw-rom}

Projection-based reduced-order modeling constructs a low-dimensional subspace
offline and evolves the system inside that subspace online, trading a controlled
loss of fidelity for a large reduction in cost. Benner et al.\ survey this
offline/online structure across parametric dynamical
systems~\cite{benner2015survey}. Simulation-oriented reduction continues to build
coordinates with physical meaning: FreeForm derives a mesh-free, material-aware
deformation basis from particle-based skinning eigenmodes to accelerate
hyperelastic simulation~\cite{xiang2026freeform}. Hybrid constructions retain
efficient reduced mechanics and learn only the part the subspace fails to resolve,
as when handle-based subspace dynamics are augmented with learned nonlinear
corrections that restore contact-induced deformation in the same reduced
coordinates~\cite{romero2021contactcorrections}. The same idea appears in vehicle
dynamics, where manifold-based linear flow-map surrogates and a learned correction
term repair a simplified vehicle model from sparse data without assuming a
prescribed form for the correction~\cite{ly2025datadriven}; that work is assessed
by the predictive accuracy of the corrected model, whereas the question here is
whether a learned NRD model is sufficient for downstream policy optimization.
The principle shared with this paper is that reduction is a decision about which
coordinates, which dynamics, and which outputs are preserved, rather than the act
of placing a smaller black-box network around the full simulator state.

Where the reduced coordinates come from separates two design attitudes. Linearly
recurrent autoencoder networks learn nonlinear latent coordinates jointly with an
approximately linear recurrence in those coordinates, so the representation is
discovered from data under a structural prior on the
dynamics~\cite{otto2019lran}. The projection used here is instead chosen from
simulator knowledge and from what the downstream controller requires. Three
consequences follow. The retained channels keep their physical identity, so the
abstraction can be inspected and revised in physical terms. The model does not
decode or reconstruct the complete Chrono state. Quantities fixed by known
relationships, including global vehicle pose and arm end-effector position, are
recovered analytically outside the network. In this paper, \emph{reduced}
therefore means discarding information that is task-irrelevant, or that the
transition does not depend on and can be recovered analytically, while preserving
the unresolved dynamics the controller depends on. It does not
mean minimizing latent dimension or reproducing every simulator output.

\subsection{High-Fidelity Simulation as the Physical Reference}
\label{subsec:rw-simulation}

All high-fidelity data in this paper are produced with Project Chrono, an
open-source multi-physics engine covering multibody dynamics, frictional
contact, finite elements, and fluid--solid
interaction~\cite{chronoOverview2016}. Two of its modules account for the two
platforms studied here. Chrono::Vehicle provides template-based ground-vehicle
models in which suspension, steering, driveline, powertrain, and tire
subsystems---and, for tracked platforms, sprocket, idler, roadwheel, and
track-shoe assemblies---are instantiated as separate components rather than
approximated by a lumped chassis model~\cite{chronoVehicle2019}. Terrain
interaction is modeled at three levels of fidelity: rigid terrain; the
semi-empirical Soil Contact Model (SCM), which supports deformable terrain for
wheeled and tracked vehicles at interactive rates~\cite{chronoSCM2019}; and the
Continuum Representation Model (CRM), an SPH-based continuum granular model
coupled to the multibody system through the FSI
interface~\cite{Huzaifa2026CRM}. Case Study~I uses rigid terrain and granular
CRM soil, with CRM defining the deformable-soil regime.

This combination is the reason Chrono is used as the reference simulator here.
Engines developed primarily for robot learning---MuJoCo~\cite{todorovMujoco2012,
mujoco_playground_2025}, Isaac Lab~\cite{isaaclab2025},
Brax~\cite{brax-freeman2021}, and Genesis~\cite{genesisSimulator}---are designed
around high-throughput vectorized rollouts, and the more recent ones add
particle-based deformable and fluid solvers, but they do not supply the vehicle
subsystem templates, powertrain and tire models, or vehicle-oriented
terramechanics that the wheeled and tracked case studies require. Chrono resolves
that physics at a throughput cost measured in Appendix~\ref{app:sim-speed}. It
is accordingly used in this work as the source of training trajectories and as
the reference in which transferred policies are evaluated, while the throughput
needed for policy search comes from the NRD models of
Sec.~\ref{sec:framework}.

\subsection{Terrain- and Context-Conditioned Dynamics}
\label{subsec:rw-context}

Off-road motion prediction improves when the learned dynamics are conditioned on
terrain. A terrain-aware kinodynamic model conditions its prediction on
terrain-related information available from proprioceptive and exteroceptive
observations, and supplies the result to a sampling-based predictive
controller~\cite{lee2023terrainaware}. Recent work sharpens this line by
propagating dynamical uncertainty several steps ahead for risk-aware speed
planning and by augmenting a nominal vehicle model with an online
Gaussian-process residual on uneven
ground~\cite{gibson2026multistep,amine2026nonplanar}. In each case the learned
component acts online, inside the control loop and continually reconciled with
incoming data, whereas the NRD models developed here are learned offline
from high-fidelity trajectories and then frozen, serving as the environment in
which a policy is trained rather than as a predictor within a planner. A
complementary group of methods dispenses with a learned dynamics model
altogether, embedding an analytical rigid-body model and an explicit rollover
constraint in GPU-parallel MPC~\cite{baxter2026highspeed} or training an
end-to-end driving policy directly inside the original
simulator~\cite{wu2026tadpo}. Several of these studies report results on
full-scale vehicles~\cite{gibson2026multistep,baxter2026highspeed,wu2026tadpo},
whereas the transfer demonstrated here returns the policy from the NRD model
to the high-fidelity simulator, in exchange for a controlled test of whether a
chosen abstraction suffices for control.

Conditioning a learned transition on a dynamics context is the general form of
this idea. Context-aware dynamics models infer a latent context from a short
history of recent transitions so that a single learned model adapts its predictions
across systems or operating conditions~\cite{lee2020contextaware}. The HMMWV study
uses the same general conditional-modeling principle with a deliberately simpler
context. The
context is a known two-class label for the interaction regime, rigid or deformable
CRM soil, rather than a latent variable inferred online; it enters the dynamics
model and not the policy observation; and it performs neither terrain recognition
nor online identification of terrain parameters. Bumpy rigid terrain accordingly
retains the rigid code, since the geometry changes while the underlying contact
regime does not. Its purpose is to resolve transition ambiguity: similar reduced
states under similar commands evolve differently under rigid and deformable
interaction.

Taken together, these lines of work fix the ingredients of the framework developed
in this paper. Learned dynamics establish that predicted experience can carry the
cost of control optimization, subject to how far the model is trusted. Reduced-order
modeling establishes that the choice of retained coordinates and outputs, not the
size of the network, determines which fidelity survives reduction. High-fidelity
simulation supplies the physical reference and the setting in which a controller is
ultimately judged. Context conditioning shows how physically distinct operating
regimes can be disambiguated when the reduced transition alone is ambiguous. The
combination addressed in this work is a task-designed physical reduction with
analytically reconstructed kinematics, a frozen model used as a fully vectorized
policy-learning environment over complete episodes, and validation by returning the
trained policy to the simulator that produced its training data. Off-road and
robot-dynamics studies generally develop a controller or a dynamics model for one
platform; the question examined here is whether deliberately selected,
task-specific abstractions can each serve as massively parallel policy-training
environments across distinct robot embodiments and control tasks.
\section{Neural Reduced Dynamics Framework}
\label{sec:framework}

Figure~\ref{fig:overview} shows the pipeline shared by both case studies.
High-fidelity Chrono trajectories are distilled into a task-specific neural
reduced dynamics (NRD) model; that model is frozen and replicated into a
vectorized environment in which a control policy is trained; and the trained
policy is returned to the full Chrono system for closed-loop validation. The
NRD model predicts a compact, task-specific reduced state rather than the complete
simulator state, and the central design decision is not the neural architecture
but how the information available from the simulator is assigned among four
roles: the recurrent state the model learns to propagate, the action applied to
the modeled system, an optional context input naming the operating regime, which
is needed when the same reduced state and action would evolve differently from
one regime to the next, and quantities recovered analytically outside the model.
This section defines that machinery; Sec.~\ref{sec:hmmwv} and
Sec.~\ref{sec:tracked} instantiate it, and Sec.~\ref{sec:cross-case} interprets
what the two instantiations teach.

\subsection{Reduced-Dynamics Formulation}
\label{subsec:framework-problem}

Let $x_t$ denote the full state of the high-fidelity Chrono system, which
advances under an applied command $a_t$ as
\begin{equation}
  \label{eq:framework-hf}
  x_{t+1} = F_{\mathrm{HF}}(x_t, a_t),
\end{equation}
where $F_{\mathrm{HF}}$ is a single Chrono step and is expensive to evaluate.
The framework does not approximate $F_{\mathrm{HF}}$ in full. It instead models
the evolution of a task-specific \emph{reduced state} $z_t = P(x_t)$, a
low-dimensional projection $P$ of the simulator state chosen per task
(Sec.~\ref{subsec:framework-architecture}). The NRD model
$f_\theta$ predicts the reduced-state increment from a finite window of the
current step and the $k$ preceding ones, using the reduced states $z_{t-k:t}$, the
applied actions $a_{t-k:t}$, and an optional context input $c_{t-k:t}$ that labels
the operating regime,
\begin{equation}
  \label{eq:framework-rom}
  \Delta \hat z_t = f_\theta\!\left(z_{t-k:t},\, a_{t-k:t},\, c_{t-k:t}\right),
  \qquad
  \hat z_{t+1} = z_t + \Delta \hat z_t,
\end{equation}
where a hat marks a model-predicted quantity, so the predicted increment $\Delta
\hat z_t$ and next state $\hat z_{t+1}$ are distinguished from the ground-truth
reduced state $z_t$. The model thus learns a residual update, and the history
supplies temporal information about actuator, contact, and terrain effects that
the instantaneous reduced state does not resolve. Equation~\eqref{eq:framework-rom}
describes teacher-forced one-step prediction during supervised training; in
open-loop rollout and policy learning, the ground-truth state history is replaced
recursively by the model-predicted history.

This formulation partitions the information available from the simulator into
four roles. The reduced state $z_t$ holds the recurrent quantities the model must
learn and propagate: the coordinates on which the transition depends and whose
evolution is uncertain or contact-dependent.
The action $a_t$ is the command actually applied to the modeled system. The
context $c_t$ is an optional regime input, used only when the same reduced state
and action evolve differently under physically distinct conditions. Finally,
quantities that are deterministic functions of the predicted state history,
written $y_t = G(\hat z_{\le t})$, are recovered analytically \emph{outside} the
model rather than assigned learned output channels. These are coordinates on
which the transition does not depend, such as global vehicle pose $(x,y,\psi)$,
integrated from the predicted body velocities, and known kinematic outputs, such
as the arm end-effector, obtained by forward kinematics on the predicted joints.

\subsection{Abstraction and Architecture}
\label{subsec:framework-architecture}

The projection $P$ follows a single rule: the reduced state retains the variables
required to propagate the task-relevant behavior, that is, the coordinates on
which the transition depends, and excludes coordinates to which the dynamics are
invariant, such as global pose on flat ground, and deterministic geometric
outputs that can be reconstructed outside the learned model. A variable is \emph{control-relevant} in this sense
when the controller would act differently if it took a different value, and the
operational test is deletion: a channel earns its place only if removing it
degrades rollout fidelity or closed-loop performance
(Appendix~\ref{app:feature-ablation}). Applying this
rule, vehicle pose, on which the planar response does not depend, is integrated
from predicted body velocities rather than predicted directly, whereas the arm's
joint positions, on which its response does depend, are retained in the state
even though they too could be integrated (Sec.~\ref{subsec:tracked-abstractions});
the arm end-effector is recovered through forward kinematics rather than learned;
and a terrain regime that changes the transition is supplied as context rather
than inferred. The two case studies show that the rule resolves
to substantially different states---a fifteen-dimensional terramechanics state
for the HMMWV, a three-dimensional planar state for the tracked vehicle, an
eight-dimensional joint-space state for the arm---without changing the principle.
The dominant physics differs across the three tasks, but the design principle is
the same: the model propagates only the coordinates on which the dynamics depend,
and everything else is supplied as an input or reconstructed from known
relationships.

All three NRD models share one architecture family: a GPT-style causal transformer
over a sequence of tokens, each concatenating the reduced state, the action, and (when
present) the context code at one time step. Causal self-attention processes a
finite window of this history, and a linear head predicts the reduced-state
increment $\Delta \hat z_t$; advancing $\hat z_{t+1} = \hat z_t + \Delta \hat
z_t$ produces the open-loop rollout. Attention over history lets the model
represent actuator- and contact-mediated effects that the instantaneous reduced
state omits and that a memoryless one-step map cannot capture. Model depth,
width, and context length are selected separately for each task from preliminary
validation and computational cost (Sec.~\ref{sec:hmmwv}, Sec.~\ref{sec:tracked}),
so the case studies share this architecture family but not an identical network
capacity or state definition.

\subsection{Training and Model Selection}
\label{subsec:framework-training}

Each NRD model is trained by supervised learning on Chrono trajectories, split at
the episode level so no trajectory is shared between training and evaluation,
with state and action channels normalized to comparable scales. The target is the
one-step reduced-state increment, trained under a per-channel weighted regression
loss
\begin{equation}
  \label{eq:framework-loss}
  \mathcal{L}_{\mathrm{step}} = \sum_j w_j\,\rho\!\left(\Delta z_{t,j} - \Delta \hat z_{t,j}\right),
\end{equation}
where $\rho$ is a squared or Huber penalty and the weights $w_j$ balance channels
whose increments differ in scale. For the tracked-vehicle and arm models the
normalized channels are already comparable, so this reduces to an unweighted
mean-squared error; the HMMWV model pools two terrain regimes whose increment
scales differ sharply, and there a per-channel domain rebalancing and the robust
Huber form keep a few high-variance channels from dominating the fit
(Sec.~\ref{subsec:hmmwv-conditioning}).

Training minimizes one-step error, but one-step error is not the quantity that
governs downstream control. A model with low one-step loss can still drift when
its own predictions are fed back over a long horizon, and it is that
autoregressive behavior the policy trains against. We nevertheless keep the
training loss one-step rather than backpropagating through multi-step
autoregressive rollouts, since gradients through a chain of the model's own
predictions are far noisier and less stable to optimize; the horizon is instead
accounted for at model selection. We therefore track two signals: a teacher-forced one-step error, and an autoregressive open-loop rollout
in which the model is advanced on recorded actions and compared to the Chrono
trajectory over a horizon. Every deployed NRD model in this paper is the checkpoint
that minimizes held-out open-loop rollout error rather than the checkpoint of
lowest one-step loss. The two signals can rank checkpoints differently, so the
choice of selection metric is consequential (Sec.~\ref{subsec:hmmwv-conditioning},
Sec.~\ref{subsec:tracked-training}).

\subsection{Policy Learning and Validation}
\label{subsec:framework-policy}

Once an NRD model is selected, its parameters $\theta$ are frozen and it becomes the
environment in which a control policy is trained. Thousands of copies of the
frozen model are batched on a single GPU and advanced in parallel, and only the
policy parameters are optimized while the learned dynamics stay fixed. During
training the policy never receives the full simulator state; its observation is
assembled from the model's own predicted reduced state $\hat z_t$ and the derived
quantities $G(\hat z_t)$, and the task reward is evaluated from the same rollout.
The frozen model advances that prediction under the resulting action,
\begin{equation}
  \label{eq:framework-policy}
  o_t = h\!\left(\hat z_t,\, G(\hat z_t)\right), \qquad a_t = \pi_\phi(o_t), \qquad
  \hat z_{t+1} = \hat z_t + f_\theta\!\left(\hat z_{t-k:t},\, a_{t-k:t},\, c_{t-k:t}\right),
\end{equation}
where $h$ is the task-specific observation map (absorbing the goal or reference,
any history, and normalization), and $a_t$ is the command finally applied to the
dynamics, with deterministic command processing folded into the policy interface.
The throughput of the frozen model---orders of magnitude above Chrono---is what
makes on-policy reinforcement learning practical
(Sec.~\ref{subsec:cross-throughput}).
Both case studies train with vectorized PPO, differing only in the observation,
reward, and action interface each task requires.

Validation proceeds at three levels of increasing stringency: 
(1) one-step accuracy, which checks the learned transition locally; 
(2) open-loop rollout, which checks whether repeated NRD predictions stay close 
to recorded Chrono trajectories; and 
(3) closed-loop transfer to Chrono, which tests whether a policy trained inside 
the frozen NRD model still works in the high-fidelity simulator. Training and this final evaluation are the same closed loop
with the same policy interface (Eq.~\ref{eq:framework-policy}), differing only in
the source of the state transition. In Chrono the trained policy is inserted
unchanged and one step reads
\begin{equation}
  \label{eq:framework-transfer}
  o_t = h\!\left(z_t,\, G(z_t)\right), \qquad a_t = \pi_\phi(o_t), \qquad
  x_{t+1} = F_{\mathrm{HF}}(x_t, a_t), \qquad z_{t+1} = P(x_{t+1}),
\end{equation}
with $z_t = P(x_t)$: the full high-fidelity state is advanced by $F_{\mathrm{HF}}$
and projected back through $P$ to rebuild the same observation. From the policy's
perspective only the dynamics provider changes: the frozen NRD model during training
and Chrono during evaluation. The policy weights, observation construction, action
processing, and safety interface carry over untouched. Closed-loop transfer is decisive because low supervised error
and high reward inside the NRD model do not by themselves rule out a policy that
exploits the model's approximation error; only evaluation in Chrono confirms that
the abstraction preserved the dynamics control requires. That test is decisive
for the abstraction, not for real-world performance: whether these policies
transfer to hardware is a question of sim-to-real transfer, which lies outside
the scope of this paper (Sec.~\ref{sec:limitations}).
Section~\ref{sec:cross-case} returns to this hierarchy and to what the two case
studies reveal about choosing the reduced state.
\section{Case Study I: Terrain-Aware HMMWV Dynamics and Control}
\label{sec:hmmwv}


This case is the strongest evidence that off-road control requires a terrain-aware
abstraction. A fixed HMMWV operates across two physically distinct terrain
regimes---flat rigid terrain and CRM deformable soil---with unseen bumpy rigid
heightmaps reserved as a strictly held-out zero-shot robustness test. Vehicle and
wheel parameters are kept the same throughout, so terrain interaction is the only source
of domain shift. All datasets, trained models, experiment configurations, and
training results reported in this paper, for both case studies, are available on the
project page~\cite{NeDMprojectPage}.

\subsection{System and Task}
\label{subsec:hmmwv-system}

The high-fidelity source is the Chrono \texttt{HMMWV\_Full} model, held fixed
across every dataset: SMC contact, a \texttt{SHAFTS} engine with an
\texttt{AUTOMATIC\_SHAFTS} transmission, all-wheel drive, and Pitman-arm
steering~\cite{chronoVehicle2019}. The task is trajectory tracking across three terrains---flat rigid
terrain, bumpy rigid heightmaps, and CRM deformable soil---that share this
vehicle and powertrain and differ only in the tire--terrain interaction. Because
the vehicle configuration is identical everywhere, terrain interaction is the
sole source of domain shift.

Data are generated with deliberately excited driver maneuvers rather than a
single path-following controller, so the training distribution spans straight-line
launch/brake, transient cornering, and combined longitudinal--lateral response.
Six maneuver families are shared across regimes sampled over low, medium, and
fast speed bands. Each episode discards an initial settling transient (a
per-scenario warmup window) before recording at 100\,Hz, and all splits are
episode-level so no trajectory is shared between training and test.

Three datasets are collected with distinct roles (Table~\ref{tab:datasets}). Flat
and CRM are used for training, validation, and in-domain testing; bumpy rigid
terrain is reserved strictly for final zero-shot out-of-distribution (OOD)
evaluation and never enters training, checkpoint selection, normalization, or
reward tuning. Bumpy terrain keeps the \emph{rigid} regime code
(Sec.~\ref{subsec:hmmwv-conditioning}): it perturbs local geometry but not the
contact physics.

\begin{table}[t]
  \centering
  \footnotesize
  \caption{HMMWV datasets and their roles. Sizes are episodes and processed
  15-D transitions (training\,/\,validation). Bumpy rigid terrain keeps the rigid
  regime code and is held out for zero-shot evaluation only.}
  \label{tab:datasets}
  \begin{tabularx}{\columnwidth}{@{}lYYc@{}}
    \toprule
    Dataset & Role & Scale & Code \\
    \midrule
    Flat rigid & Train, val, in-domain test & $\approx$82k episodes ($\sim$300\,GiB); 329\,M\,/\,81\,M transitions & $[1,0]$ \\
    CRM soil   & Train, val, in-domain test & 2{,}000 episodes; 2.28\,M\,/\,0.60\,M transitions & $[0,1]$ \\
    Bumpy rigid & Zero-shot OOD test only & 20 episodes are selected for chrono closed-loop policy evaluation & $[1,0]$ \\
    \bottomrule
  \end{tabularx}
\end{table}

The three regimes differ substantially in their contact physics and,
correspondingly, their simulation cost (Table~\ref{tab:sim-settings}). Flat and
bumpy rigid terrain use \texttt{TMEASY} tires~\cite{Rill15} on an SMC rigid contact surface at
friction $\mu=0.9$, integrated at a 2\,ms step. The CRM terrain replaces the rigid surface with a Chrono SPH--FSI deformable-soil model~\cite{weiGranularSPH2021,Huzaifa2026CRM}:
the four wheels are registered as FSI rigid bodies carrying rigid
tires, spindle forces are read from the FSI solver rather than an analytical tire
model, and the soil is a cohesive granular medium (density
$1700$\,kg/m\textsuperscript{3}, cohesion $5$\,kPa, internal friction $0.8$,
$E=1$\,MPa). The SPH field uses $0.08$\,m initial particle spacing over a
$150{\times}150{\times}0.25$\,m box with a moving active domain, and requires a
$0.5$\,ms integration step---$4\times$ finer than rigid---on top of a per-step
neighbor search, making CRM episodes markedly more expensive to simulate than
rigid ones. Appendix~\ref{app:sim-speed} reports the measured throughput of each
regime and of the neural surrogate.

\begin{table}[t]
  \centering
  \footnotesize
  \caption{Simulation settings by terrain regime. All regimes share the fixed
  \texttt{HMMWV\_Full} vehicle (SMC contact, \texttt{SHAFTS} engine, automatic
  transmission, AWD, Pitman-arm steering) and record at 100\,Hz; they differ only
  in the tire--terrain contact model.}
  \label{tab:sim-settings}
  \begin{tabularx}{\columnwidth}{@{}lYY@{}}
    \toprule
    Setting & Rigid (flat, bumpy) & CRM (soil) \\
    \midrule
    Tire model        & \texttt{TMEASY} (analytical) & \texttt{RIGID\_MESH} (FSI) \\
    Terrain / contact & SMC rigid surface & SPH--FSI deformable soil \\
    Tire-force source & analytical tire report & FSI spindle force \\
    Friction / soil   & $\mu=0.9$ & coh.\ $5$\,kPa, $\mu_s{=}0.8$, $\rho{=}1700$\,kg/m\textsuperscript{3}, $E{=}1$\,MPa \\
    Integration step  & 2\,ms & 0.5\,ms \\
    Terrain extent    & flat / $500{\times}500$\,m heightmap ($\pm0.6$\,m) & $150{\times}150{\times}0.25$\,m box, $0.08$\,m spacing \\
    \bottomrule
  \end{tabularx}
\end{table}

The maneuver mix is preserved across regimes but adapted to CRM's finite domain:
the CRM set uses shorter $12$--$18$\,s episodes and reduced top speed and steering
amplitude so that turning trajectories stay inside the $150$\,m box. Bumpy rigid terrain is
generated from a heightmap library (\texttt{bumpy\_field\_*} patches,
$500{\times}500$\,m, height amplitude $\pm0.6$\,m) applied to the same rigid
contact model. Its held-out test set covers the same six maneuver families, and
every closed-loop evaluation is run on the exact heightmap its reference
trajectory was recorded on, so the desired path and the terrain beneath it always
correspond.

\subsection{Reduced State Abstraction}
\label{subsec:hmmwv-state}

\begin{figure}[t]
  \centering
  \includegraphics[width=\textwidth]{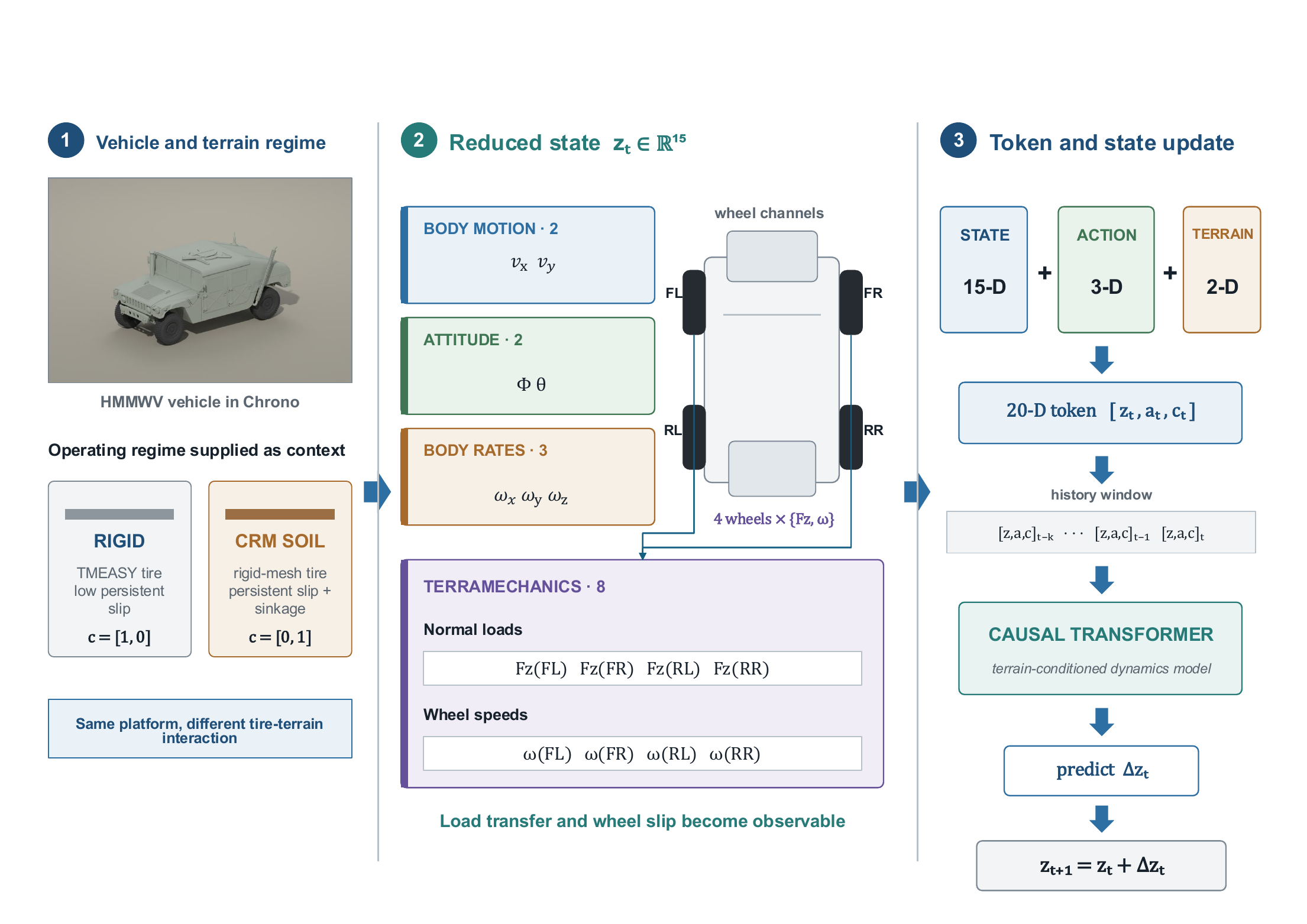}
  \caption{Control-relevant reduced state for the HMMWV. A fixed vehicle interacts
  with rigid terrain ($c=[1,0]$) or CRM deformable soil ($c=[0,1]$); the $15$-D
  reduced state $z_t$ collects body motion ($v_x,v_y$), attitude ($\phi,\theta$),
  body-frame rates ($\omega_x,\omega_y,\omega_z$), and the eight-channel
  terramechanics block of tire normal loads $F_z^{(i)}$ and wheel speeds
  $\omega^{(i)}$ that make slip and load transfer observable. At each step the state,
  the $3$-D action, and the $2$-D terrain code form a $20$-D token; the causal
  transformer NRD model predicts the increment $\Delta z_t$ and advances
  $z_{t+1}=z_t+\Delta z_t$.}
  \label{fig:state-diagram}
\end{figure}

The reduced-state design uses the following 15-dimensional state:
\begin{equation}
  \small
  \label{eq:state-15d}
  z =
  \begin{bmatrix}
    v_x,\; v_y, & \text{(body longitudinal/lateral velocity)}\\
    \phi,\; \theta, & \text{(roll, pitch)}\\
    \omega_x,\; \omega_y,\; \omega_z, & \text{(body-frame roll/pitch/yaw rates)}\\
    F_z^{\mathrm{fl}},\,F_z^{\mathrm{fr}},\,F_z^{\mathrm{rl}},\,F_z^{\mathrm{rr}}, & \text{(tire normal forces)}\\
    \omega^{\mathrm{fl}},\,\omega^{\mathrm{fr}},\,\omega^{\mathrm{rl}},\,\omega^{\mathrm{rr}} & \text{(wheel angular velocities)}
  \end{bmatrix}.
\end{equation}
The action is $a = [\,\text{steering},\ \text{throttle},\ \text{braking}\,]$, the
three driver commands. The first seven channels---planar body velocities
$v_x,v_y$, attitude $\phi,\theta$, and the roll/pitch/yaw rates---are the
body-motion base. The remaining eight are the terramechanics block: the four tire normal (vertical)
loads $F_z^{(i)}$ and the four wheel angular velocities $\omega^{(i)}$, one per
wheel $i\in\{\mathrm{fl},\mathrm{fr},\mathrm{rl},\mathrm{rr}\}$.

The eight terramechanics channels make the reduced state responsive to terrain
interaction by exposing load redistribution and wheel--body motion mismatch. The normal
loads $F_z^{(i)}$ carry the load transfer produced by acceleration, braking, and
cornering, and on soft or uneven terrain they additionally carry information about
the contact-load redistribution induced by deformable or uneven ground that a
rigid-terrain model never observes. The wheel
speeds $\omega^{(i)}$ play the analogous role for the drive axis: together with the
body velocity they define an illustrative per-wheel longitudinal-slip proxy,
\begin{equation}
  \label{eq:slip}
  s^{(i)} = \frac{\omega^{(i)} r - v_x}{\max\!\left(|v_x|,\ \epsilon\right)},
\end{equation}
where $r$ is the wheel rolling radius. On rigid terrain the wheel rolls almost
freely, $v_x \approx \omega^{(i)} r$ and $s^{(i)} \approx 0$; on deformable CRM
soil the same drive torque produces persistent slip of order $25\%$ together with
sinkage. Slip is thus the primary quantity that separates the two regimes, and
Eq.~\eqref{eq:slip} makes explicit that it is representable only if the reduced
state carries \emph{both} the wheel speed and the body velocity.
Absolute position $(x,y)$ and heading $\psi$ are excluded from $z$, since the
vehicle response on the training terrains does not depend on them, and are
reconstructed during rollout by integrating the predicted body velocities and yaw
rate (Sec.~\ref{subsec:framework-problem}); this keeps the learned state compact
and invariant to where and in which direction an episode starts.

We adopt this 15-D state as a deliberate, compact design choice: the body-motion
base captures vehicle handling, and the four normal loads and four wheel speeds add
exactly the terramechanical quantities---load transfer and wheel slip---through
which terrain shapes the response. The state-abstraction ablation of
Appendix~\ref{app:feature-ablation} confirms this choice: dropping the four
normal forces and four wheel speeds from the state and prediction target costs
little on rigid terrain but degrades open-loop CRM rollout error by $\sim$$49\%$.

\subsection{Terrain-Conditioned NRD Model}
\label{subsec:hmmwv-conditioning}

This subsection defines the terrain-conditioned dynamics model at the core of the
HMMWV case. Terrain conditioning is needed because
the same reduced state--action pair evolves differently on rigid terrain and on CRM
deformable soil. Under comparable driving conditions, CRM exhibits substantially
larger and more persistent wheel slip as well as a noisier state-transition process. Naively mixing the rigid and CRM training data therefore would average this domain difference and degrade the model performance. We resolve
this ambiguity with an explicit two-class terrain code: rigid is $c=[1,0]$ and CRM
is $c=[0,1]$, repeated at every step and concatenated with the state and action so
that each token grows from $18$ to $20$ dimensions. Concatenating a small one-hot
context code is a standard, low-cost mechanism that lets one shared backbone reuse
the common vehicle-dynamics representation while specializing the domain shift between rigid and CRM transitions.
Removing the terrain key (Appendix~\ref{app:feature-ablation}) leaves one-step loss
almost unchanged but more than doubles open-loop rollout error on flat terrain,
confirming that its value shows up in compounding rollout error rather than
one-step prediction.

\paragraph{Architecture.} The NRD model $f_\theta$ instantiates the
shared causal-transformer backbone of Sec.~\ref{subsec:framework-architecture} for
the HMMWV. Each input token concatenates the 15-D reduced state, the 3-D action, and
the 2-D terrain code---$20$ channels in total---and the model attends over a context
of up to $128$ steps to predict the 15-D state increment $\Delta z$. The transformer
uses $8$ layers, $8$ attention heads, and a $256$-dimensional embedding (no bias,
with a $256$-unit prediction head), totaling $6.40$\,M parameters; it remains
deliberately compact relative to frontier-scale transformers, since the model's
value is throughput rather than raw capacity. This depth is not an arbitrary choice:
Appendix~\ref{app:ablation-ofat} reports an architecture sweep over depth, width,
attention heads, and context length, and finds depth the dominant lever, with
returns saturating beyond depth $8$. All results in this section use the resulting
$8$-layer backbone.

\paragraph{Training.} Training one model across both terrains is not
simply a matter of pooling the data. A flat-terrain specialist does not transfer to
CRM at all; sequential finetuning from the flat model onto CRM data degrades the
previously learned rigid behavior; and naively mixing the two under a plain
mean-squared error is dominated by CRM's tire-force channels, whose per-step
increments are far larger than the rigid ones, so both the optimizer and the
checkpoint metric chase an essentially aleatoric signal. The corrected procedure
draws flat and CRM sub-batches at a fixed $75/25$ ratio, attaches the terrain code,
normalizes inputs consistently, and replaces the plain loss with a per-channel
domain-rebalanced Huber loss
\begin{equation}
  \label{eq:cotraining-loss}
  \mathcal{L} = \sum_j w_j\,\mathrm{Huber}\!\left(\Delta z_j - \widehat{\Delta z}_j\right),
\end{equation}
whose weights $w_j$ derive from the combined flat--CRM per-channel scale, so the
high-variance CRM force deltas no longer overwhelm the lower-scale but
control-critical channels. Checkpoints are selected on long-horizon rollout rather
than one-step loss: each epoch, the model is rolled open-loop over held-out flat and
CRM episodes and selection minimizes the domain-balanced rollout error
\begin{equation}
  \label{eq:worst-domain}
  S(\theta) = \tfrac{1}{2}\,E_{\mathrm{rigid}}(\theta) + \tfrac{1}{2}\,E_{\mathrm{CRM}}(\theta),
\end{equation}
where $E_d$ is the open-loop position error on domain $d$ normalized by the distance
traveled, which makes the shorter, slower CRM episodes comparable to rigid ones. The
per-domain errors are reported alongside the scalar so that neither terrain can
dominate selection unnoticed.

\begin{figure}[t]
  \centering
  \includegraphics[width=0.95\textwidth]{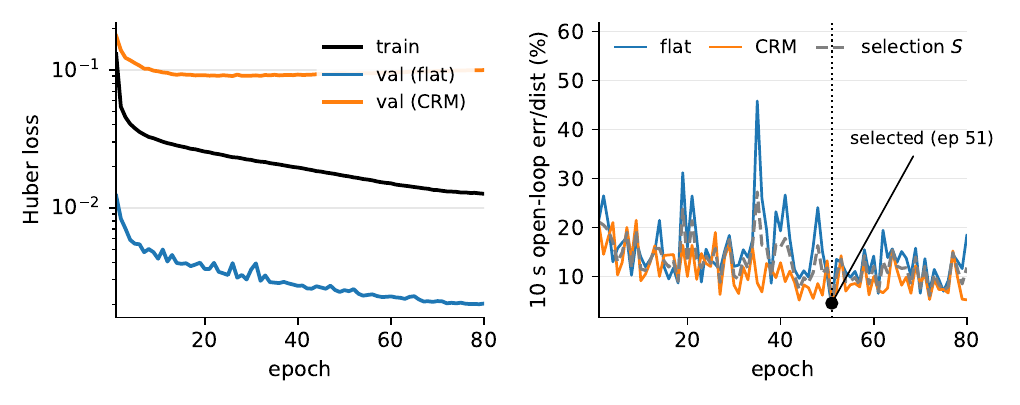}
  \caption{Training the terrain-conditioned $8$-layer generalist (flat$+$CRM
  co-training). \emph{Left}: training and per-domain validation one-step Huber loss;
  the CRM validation loss plateaus higher than the flat case given its larger variation during dynamics transitions. \emph{Right}:
  $10$\,s open-loop rollout error normalized by distance for each domain and the
  domain-balanced selection score $S$ (Eq.~\ref{eq:worst-domain}); the epoch-$51$
  checkpoint that minimizes $S$ is selected.}
  \label{fig:training}
\end{figure}

The model is optimized with AdamW -- $\beta=(0.9,0.95)$, weight
decay $0.1$ -- at a peak learning rate of $3\times10^{-4}$ decayed to $3\times10^{-5}$
after a $1000$-step warmup, with gradients clipped to unit norm. Each of $80$ epochs
draws $2000$ mini-batches of $64$ history windows at the $75/25$ flat/CRM ratio
($1.6\times10^{5}$ updates in total); training the full conditioned $8$-layer
generalist takes about $85$ minutes on a single GPU (NVIDIA RTX 4090). The final
baseline is trained using the full designated training split:
Appendix~\ref{app:data-scaling} shows
rollout accuracy keeps improving as more training episodes are added. This observation aligns with transformer-based
scaling laws in other robotics applications, which makes it promising for scaling-up
and generalizability. 
Figure~\ref{fig:training} shows why checkpoints
are selected by rollout rather than loss: the one-step validation loss keeps
falling, but on CRM it is dominated by the near-aleatoric tire-force channels and
tracks rollout quality poorly, whereas the $10$\,s open-loop rollout error---the
quantity scored by Eq.~\eqref{eq:worst-domain}---identifies the epoch-$51$
checkpoint used in all downstream evaluation.

\subsection{Dynamics Model Evaluation}
\label{subsec:hmmwv-dynamics-eval}

The purpose of this evaluation is to test whether a single learned dynamics model
can serve both terrain regimes, and in doing so to show why training on either
terrain alone is insufficient. The model is judged primarily by long-horizon
open-loop rollout error normalized by distance traveled, which is far more
informative for downstream policy learning than one-step validation loss.
Table~\ref{tab:terrain-conditioned} compares the co-trained terrain-conditioned
generalist against two single-terrain specialists---one trained only on rigid data,
one only on CRM---on held-out rigid and CRM episodes.

\begin{table}[t]
  \centering
  
  \footnotesize
  \caption{Terrain-conditioned generalist vs.\ single-terrain specialists
  (rigid-only, CRM-only), each at its selected checkpoint. \emph{One-step loss} is
  the channel-reweighted Huber validation loss; \emph{rollout err/dist} is the
  10\,s open-loop position error normalized by distance traveled.
  \textsuperscript{\dag}\,zero-shot: specialist evaluated on the terrain it was not
  trained on.}
  \label{tab:terrain-conditioned}
  \setlength{\tabcolsep}{9pt}
  \begin{tabular}{@{}l r r r r@{}}
    \toprule
     & \multicolumn{2}{c}{One-step loss} & \multicolumn{2}{c}{10\,s rollout err/dist} \\
    \cmidrule(lr){2-3}\cmidrule(lr){4-5}
    Dynamics model & Rigid & CRM & Rigid & CRM \\
    \midrule
    Generalist (75/25) & 0.0025 & 0.094 & 3.7\% & 5.4\% \\
    Rigid-only         & 0.0018 & 1.27\textsuperscript{\dag}  & 3.8\% & 194.1\%\textsuperscript{\dag} \\
    CRM-only           & 0.72\textsuperscript{\dag} & 0.11 & 39.6\%\textsuperscript{\dag} & 4.2\% \\
    \bottomrule
  \end{tabular}
\end{table}

The result shows why co-training is necessary, and it shows something stronger than
a robustness trade-off. Each single-terrain specialist is close to the generalist on
its own terrain (rigid-only $3.8\%$, CRM-only $4.2\%$) but collapses off it---the
rigid-only model reaches $194\%$ error on CRM and the CRM-only model $40\%$ on
rigid---so neither is usable across regimes. The generalist remains close to the respective
specialist on each in-domain test ($3.7\%$ rigid, $5.4\%$ CRM) while avoiding
catastrophic cross-domain failure.

These are open-loop rollout errors on held-out episodes; closed-loop transfer of a
single policy trained inside this conditioned model is evaluated in Chrono in
Sec.~\ref{subsec:hmmwv-policy-transfer}.

\subsection{RL Tracking and Terrain-Conditioned Policy Transfer}
\label{subsec:hmmwv-rl}
\label{subsec:hmmwv-policy-transfer}

With the conditioned dynamics model frozen, we train a trajectory-tracking policy
entirely with it. The policy is
optimized with PPO across $2{,}048$ vectorized copies of the frozen model on a
single GPU; the actor and critic are $512$--$256$--$128$ ELU multilayer perceptrons
with empirical observation normalization. At each control step the policy receives
the $231$-dimensional observation
\begin{equation}
  \label{eq:hmmwv-obs}
  o_t =
  \begin{bmatrix}
    z_{\tau-9:\tau}                            & \text{state history ($10\times15$)}\\
    a_{\tau-10:\tau-1}                         & \text{action history($10\times3$)}\\
    z_\tau - z_\tau^{\mathrm{ref}}             & \text{state error ($15$)}\\
    e^{\mathrm{pose}}_t                        & \text{body-frame error ($3$)}\\
    p_{t+1:t+10}                               & \text{reference preview ($10\times3$)}\\
    a_{t-1}                                    & \text{last action ($3$)}
  \end{bmatrix}
  \in \mathbb{R}^{231},
\end{equation}
where $z_{\tau-9:\tau}$ and $a_{\tau-10:\tau-1}$ denote the $10$-step
reduced-state and action histories, $z_\tau-z_\tau^{\mathrm{ref}}$ is the
reduced-state tracking error, $e^{\mathrm{pose}}_t$ is the
longitudinal/lateral/heading pose error in the body frame, and
$p_{t+1:t+10}$ are the upcoming reference poses, each $p$ contains reference $(x, y)$ coordinate and heading angle; all entries are normalized, and the blocks sum to
$231$. The observation deliberately excludes the terrain code:
terrain enters only through the frozen conditioned dynamics that generate the
rollout, and implicitly through the tire-force and wheel-speed state channels, so a
single control policy is learned across regimes. The policy outputs the three driver
commands $a=[\delta,\tau,\beta]$ (steering, throttle, brake), clipped to their
physical ranges with the steering channel additionally rate-limited between
consecutive steps (to prevent solver failures due to unsmoothed steering change). One command is held for five dynamics steps, giving $20$\,Hz
control over the $100$\,Hz reduced dynamics ($\sim$$9$\,s episodes). Training draws on
$40$ reference segments---$20$ flat and $20$ CRM, each a random $\sim$$11$\,s
mid-episode window covering the six maneuver families; at every episode reset the
environment randomly samples one reference matched to that environment's terrain and
warm-starts the rollout from the segment's initial recorded history. The RL
environment queries the frozen dynamics model with only the last $k{=}16$ of its
$128$-step training context (Sec.~\ref{subsec:hmmwv-conditioning}). This is faster:
attention cost is quadratic in context length, and the shorter window is what
sustains the $\approx 35{,}000$ policy-control steps/s aggregate throughput
($\approx 6.8\times$ faster than the $\approx 5{,}080$ steps/s measured at full
$128$-step context, same GPU and config) used to train all three policies
(Appendix~\ref{app:sim-speed}). It comes at no cost to
tracking quality: an open-loop truncation sweep shows $300$-step pose RMSE flat
across the full context range, and policies trained under $k{=}16$ and
full-context dynamics reach the same median closed-loop tracking error. 

Closed-loop
\emph{evaluation} instead uses a separate held-out set of $20$ rest-start references
per terrain, whose segments begin at vehicle rest so Chrono can warm-start each
rollout from zero speed at the reference pose. Note that the closed-loop evaluation trajectories are seen by neither the dynamics model nor the RL policy during training.  

\begin{figure}[t]
  \centering
  \includegraphics[width=0.6\textwidth]{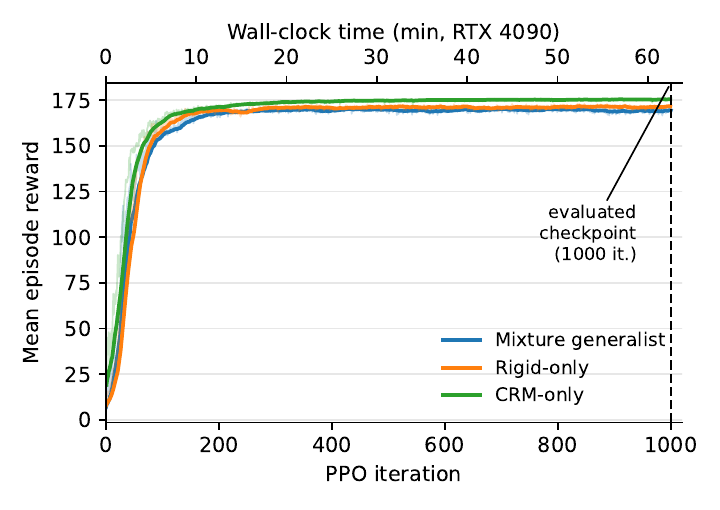}
  \caption{PPO training convergence for the three tracking
  policies (mixture generalist and rigid-only / CRM-only specialists), each trained
  inside the frozen $8$-layer conditioned NRD model through $1000$ iterations. The top axis converts PPO iteration to wall-clock
  training time on a single RTX~4090; faint traces are raw per-iteration values; solid curves are
  exponentially smoothed.}
  \label{fig:rl-reward}
\end{figure}

The per-step reward pairs a Gaussian tracking term with two small action penalties,
\begin{equation}
  \label{eq:hmmwv-reward}
  r_t = \exp\!\big(-\mathcal{L}_{\mathrm{track}}\big)
        \;-\; \lambda_{a}\,\lVert a_t - a_{t-1}\rVert^2
        \;-\; \lambda_{tb}\,\tau_t\,\beta_t,
\end{equation}
whose tracking loss aggregates the position, heading, and body-state errors to the
reference,
\begin{equation}
  \label{eq:hmmwv-track-loss}
  \mathcal{L}_{\mathrm{track}} =
    w_p\!\left(\frac{e_p}{\sigma_p}\right)^{\!2}
    + w_\psi\!\left(\frac{e_\psi}{\sigma_\psi}\right)^{\!2}
    + \frac{w_s}{|\mathcal{S}|}\sum_{j\in\mathcal{S}}
      \left(\frac{z_j-z_j^{\mathrm{ref}}}{\sigma_s\,\hat{\sigma}_j}\right)^{\!2}.
\end{equation}
Here $e_p$ and $e_\psi$ are the position and yaw errors to the reference,
$\mathcal{S}=\{v_x,v_y,\omega_z\}$ are the tracked body-state channels scaled by their
dataset standard deviations $\hat{\sigma}_j$, and the $\tau_t\beta_t$ term discourages
simultaneous throttle and brake. We use $(w_p,w_\psi,w_s)=(2.0,1.6,0.2)$,
$(\sigma_p,\sigma_\psi,\sigma_s)=(2.0\,\mathrm{m},\,0.35\,\mathrm{rad},\,1.0)$, and
$(\lambda_a,\lambda_{tb})=(0.2,0.05)$; episodes terminate when the position error
exceeds $1$\,m or the roll/pitch bounds are exceeded. The mixture generalist and the
two single-terrain specialists used below share this setup and converge well within
budget: the mean episode reward rises steeply and plateaus by $\sim$$300$ iterations
(Fig.~\ref{fig:rl-reward}), so we evaluate the $1000$-iteration checkpoint of each.

\paragraph{Single-terrain specialists.} For comparison we train two specialist
policies that follow the \emph{identical} recipe---the same observation
(Eq.~\ref{eq:hmmwv-obs}), reward (Eqs.~\ref{eq:hmmwv-reward}--\ref{eq:hmmwv-track-loss}),
PPO architecture, context length $k{=}16$, and action clipping---and differ only in the
frozen dynamics model and terrain-specific training data. The rigid-only policy is
trained with the rigid-only dynamics model (the rigid-only specialist of
Sec.~\ref{subsec:hmmwv-dynamics-eval}) on $20$ flat references with the terrain code
held to rigid; the CRM-only policy is trained with the CRM-only dynamics model on
$20$ CRM references with the code held to CRM. Because the PPO architecture, reward,
observation design, and action constraints are held fixed, differences in transfer
primarily reflect the learned training dynamics and terrain coverage rather than
changes to the controller design.

\begin{figure}[t]
  \centering
  \includegraphics[width=0.6\textwidth]{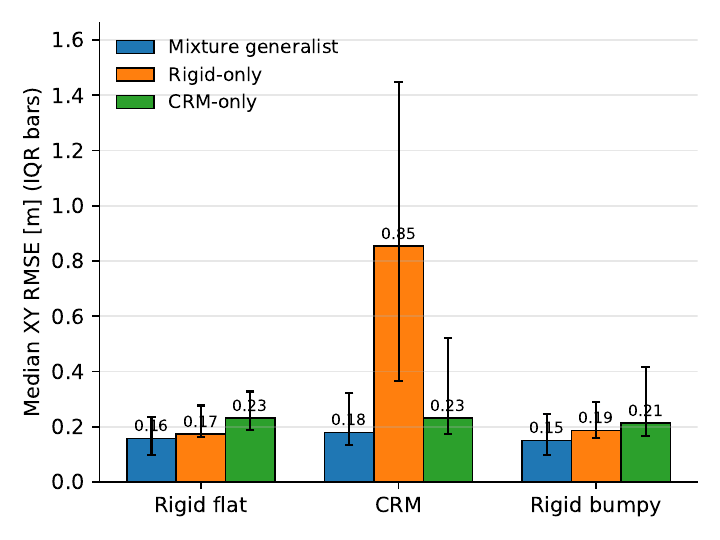}
  \caption{Closed-loop Chrono policy transfer across the three terrain regimes: one
  mixture generalist vs.\ the rigid-only and CRM-only
  specialists, all trained on the $8$-layer dynamics backbone
  (Sec.~\ref{subsec:hmmwv-conditioning}, Appendix~\ref{app:ablation-ofat}) and
  evaluated at their $1000$-iteration
  checkpoint under the same action clipping. Bars are the median XY RMSE over $20$
  held-out references; whiskers span the interquartile range. The mixture
  generalist achieves the lowest median and mean on all three evaluated terrains,
  whereas each specialist degrades sharply off its training terrain---most visibly the
  rigid-only policy on CRM. Bumpy is a strict zero-shot test (rigid code, no bumpy
  data in training or selection). Full per-terrain statistics are given in
  Table~\ref{tab:policy-transfer}, Appendix~\ref{app:policy-transfer}.}
  \label{fig:policy-transfer-bars}
\end{figure}

\paragraph{Three-terrain transfer.} The decisive test is closed-loop transfer of a
\emph{single} policy, trained inside the terrain-conditioned dynamics model, back
to Chrono across all three regimes.
Figure~\ref{fig:policy-transfer-bars} compares the mixture generalist---one policy
trained with the rigid/CRM mixed dynamics model (Sec.~\ref{subsec:hmmwv-conditioning})---against the two single-terrain
specialists, on held-out rigid flat, CRM, and zero-shot bumpy references under the
same action clipping. All nine policy$\times$terrain cells complete $20/20$ rollouts
with zero early terminations (full per-terrain statistics---median, mean, and
Interquartile Range (IQR)---in Table~\ref{tab:policy-transfer}, Appendix~\ref{app:policy-transfer}). The generalist policy
\emph{achieves the lowest mean and median XY RMSE on all three evaluated terrains}.
It beats the rigid-only specialist on rigid flat itself ($0.157$ vs.\ $0.174$\,m
median) and beats the CRM-only specialist on CRM itself ($0.180$ vs.\ $0.231$\,m
median), in addition to its expected wins on bumpy ($0.149$ vs.\ $0.187$/$0.213$\,m
median). Each specialist, by contrast, degrades sharply off its training terrain:
the rigid-only policy collapses on CRM (median $0.854$\,m, mean $1.000$\,m) and the
CRM-only policy is worst on both rigid regimes, see Fig.~\ref{fig:policy-traj}. This
is the closed-loop counterpart to the dynamics-level result of
Sec.~\ref{subsec:hmmwv-dynamics-eval}: with the selected $8$-layer backbone, the
generalist policy attains lower median and mean tracking error than both specialist
policies on every evaluated terrain.

The bumpy column is a strict zero-shot test: no bumpy data enters training,
checkpoint selection, normalization, or reward tuning, and bumpy terrain carries the
same rigid code as flat terrain (Sec.~\ref{subsec:hmmwv-system}). Because the policy
never previews the height profile, closed-loop Chrono tracking is the honest metric,
and the generalist's bumpy median ($0.149$\,m), the lowest among the compared
policies, shows that a policy
trained only on flat and CRM data maintains low tracking error on unseen rough
rigid terrain---evidence that the conditioned abstraction generalizes rather than
memorizes.
\section{Case Study II: Reduced Dynamics for a Tracked Vehicle With an Articulated Arm}
\label{sec:tracked}

Case Study~I asked whether a suitable reduced state choice can preserve
terrain-dependent vehicle dynamics across substantially different contact
regimes. Case Study~II probes the translational dimension of the methodology presented here by asking a complementary question: can the \emph{same}
``high-fidelity-data''--to--``policy'' pipeline be reused across different robotic
subsystems and control objectives in the absence of a common full-system state? To address this question, we consider a Chrono tracked vehicle with a
front-mounted articulated manipulator arm that is subjected to two independent control tasks:
planar goal-reaching for the tracked vehicle, and end-effector goal-reaching for
the arm. Each task is served by its own task-specific NRD model;
each policy is trained entirely inside the corresponding frozen NRD model and then
validated in the full Chrono system. The two abstractions differ substantially
in dimension and physical meaning, yet both instantiate the causal next-state
prediction of Sec.~\ref{sec:framework}. We do \emph{not} present this as
simultaneous locomanipulation or coordinated vehicle--arm control; the point is
that the right abstraction changes with the subsystem and the task.

\subsection{Purpose and Experimental Design}
\label{subsec:tracked-purpose}
\begin{figure}[t]
  \centering
  \includegraphics[width=0.85\textwidth]{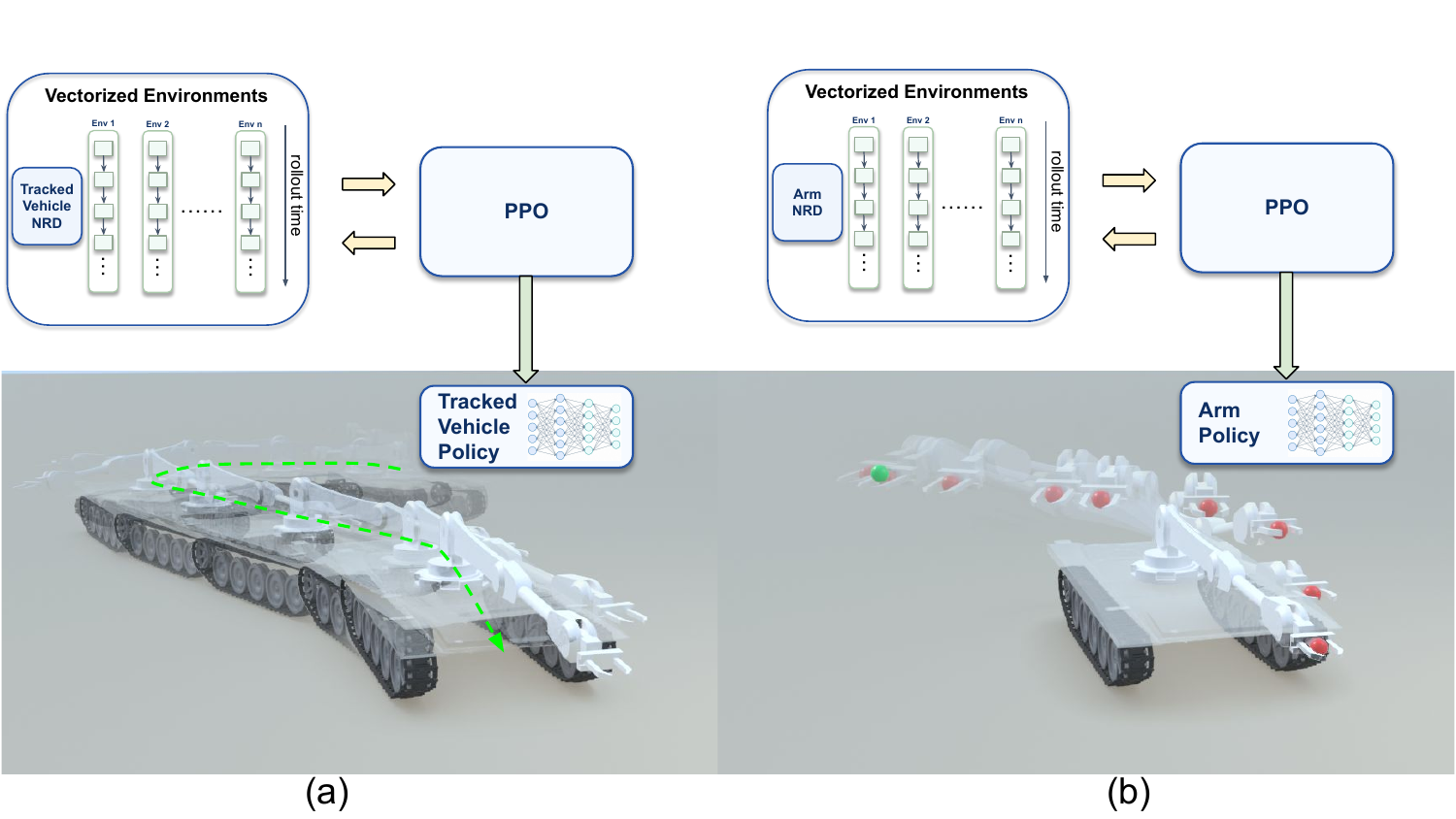}
  \caption{Platform and two-task decomposition for Case Study~II on one
  vehicle-plus-arm system. (a)~Goal reach: the tracked vehicle maneuvers to a planar
  goal while the front-mounted arm rides at its home configuration. (b)~Arm-case
  overview: with the vehicle held stationary, the $4$-DOF arm positions its
  end-effector at targets across its workspace. Each task has its own reduced
  state, learned dynamics model, and policy, and each is validated back in
  high-fidelity Chrono.}
  \label{fig:tracked-platform}
\end{figure}

The platform is a stock single-pin tracked vehicle from Chrono::Vehicle (the
\texttt{M113} model, with 154 rigid bodies and more than 150 degrees of freedom) with a $4$-DOF gripper arm rigidly mounted at the front. The two tasks exercise the platform in two modes. In \emph{drive mode}, the vehicle maneuvers to a planar goal
while the arm is held at its home configuration as fixed mounted mass, so the
NRD model must capture the vehicle's response to driver commands including
the mass and center-of-gravity offset the arm contributes. In \emph{reach mode},
the vehicle is stopped and only the arm moves, so the NRD model must capture the
joint response under command-driven actuation, from which end-effector motion is
recovered through forward kinematics. Data for both
modes are collected in this same mounted-arm Chrono scene, recording at each
control step the reduced state and the command applied. In drive mode the
welded-arm vehicle is excited by ten randomized maneuver families---steering steps,
doublets, sine and chirp sweeps, pivot-like turns, launch--brake,
steer-while-braking, multi-steer, coast-down, and stop-and-go---whose amplitudes,
frequencies, and durations are drawn per episode to span the vehicle's operating
envelope. In reach mode the arm follows a smooth-command sampler that draws a
random joint target within the limits and advances $q^{\mathrm{cmd}}$ toward it at
a bounded rate (resampled about once per second), giving low-jerk actuation rather
than white noise; collection is restricted to free-space motion, each episode
monitoring arm--ground, arm--vehicle, and arm--self contact by a forward-kinematics
signed-distance test (Appendix~\ref{app:arm-safety}) and terminating on contact so
no in-collision transitions corrupt the learned distribution.
Figure~\ref{fig:tracked-platform} summarizes the platform and the two-task
decomposition that organizes the rest of this section. Together, the two modes
exercise the complementary capabilities of such a platform: driving the vehicle to a
target location and positioning the arm end-effector at a target within its
workspace. Developing controllers for such a system directly is computationally
demanding---the high-fidelity tracked-vehicle scene carries hundreds of degrees of freedom
and rich terramechanics interactions, so every simulation step is costly and
full-system policy search is impractical. Routing each mode through the
neural reduced dynamics pipeline is what makes the problem tractable: it lets us
design and iterate the reinforcement-learning solution efficiently inside compact
learned dynamics. Concretely, the high-fidelity vehicle-plus-arm scene steps at only
$0.28\times$ real time, whereas the NRD models roll out three to four orders
of magnitude faster---up to $58{,}400\times$ cheaper per simulated second
(Appendix~\ref{app:sim-speed-tracked}).

\subsection{Task-Specific State Abstractions and NRD Models}
\label{subsec:tracked-abstractions}

\paragraph{Planar tracked-vehicle abstraction.} The drive-mode reduced state keeps
only the planar body motion needed to propagate vehicle pose,
\begin{equation}
  \label{eq:tracked-vehicle-state}
  z^{\mathrm{veh}} = [\,v_x,\ v_y,\ r\,] \in \R^{3},
\end{equation}
the body-frame longitudinal and lateral velocities and the yaw rate, with action
$a^{\mathrm{veh}} = [\,\text{steering},\ \text{throttle},\ \text{braking}\,]$.
The global position $(x,y)$ and heading $\psi$, on which the planar response
does not depend, are excluded from the learned state and reconstructed during
rollout by integrating the predicted body velocities and yaw rate
(Sec.~\ref{subsec:framework-problem}). We keep only these
channels because they suffice for the control task: $z^{\mathrm{veh}}$ carries
everything the goal-reaching policy needs to be trained in this application.
The reduction is aggressive: the Chrono vehicle advances
154 rigid bodies and more than 150 degrees of freedom every step, among them two
track chains with their sprocket, idler, road-wheel, and track-shoe assemblies
(Sec.~\ref{subsec:tracked-purpose}), while the learned model propagates three
numbers. None of that mechanism appears in $z^{\mathrm{veh}}$; what survives is
the planar motion it produces, which is all the goal-reaching controller acts
through.

\paragraph{Articulated-arm abstraction.} The reach-mode reduced state is
\begin{equation}
  \label{eq:tracked-arm-state}
  z^{\mathrm{arm}} = [\,q,\ \dot q\,] \in \R^{8},
\end{equation}
the four joint positions $q$ and velocities $\dot q$. The action is the four joint
command targets $a^{\mathrm{arm}} = q^{\mathrm{cmd}} \in \R^{4}$, the setpoint the
actuators are driven toward over the step.

First, the arm is commanded through $q^{\mathrm{cmd}}$ under a per-joint PD
impedance control rather than through raw joint torque. PD impedance control is
deterministic and yields smoother joint motion than direct torque commands, which
gives better-conditioned transitions for the NRD model to learn. Because
$q^{\mathrm{cmd}}$ is an exogenous setpoint the actuators are regulated toward---an
input to the dynamics, not an internal state that evolves on its own---we treat it
as the \emph{action} rather than as a state channel. The reaching policy still
moves the setpoint by bounded increments $\Delta q^{\mathrm{cmd}}$ for smooth
actuation (Sec.~\ref{subsec:tracked-policies}), but that increment is a property of
the policy, not a dimension the dynamics model must carry.

Second, we \emph{exclude} the end-effector from the reduced state and recover its
position by forward kinematics on the predicted joints,
$p^{\mathrm{ee}} = \mathrm{FK}(q)$. The end-effector is a deterministic geometric
function of $q$, which the model already predicts, so it needs no separate learned
channel. Forward kinematics is in any case evaluated every step inside the
rollout---the collision-safety shield (Appendix~\ref{app:arm-safety}) already
re-derives the end-effector and link geometry from $q$---so recovering
$p^{\mathrm{ee}} = \mathrm{FK}(q)$ costs nothing extra and keeps the reaching target
exactly on the kinematics. These two choices leave a compact eight-dimensional
reduced state, $[q,\dot q]$, that we found accurate enough for open-loop rollout and
downstream reaching (Sec.~\ref{subsec:tracked-training}).

The joint positions are retained rather than integrated from the predicted
velocities, as vehicle pose is, because the model should receive the coordinates
on which the one-step dynamics depend (Sec.~\ref{subsec:framework-architecture}).
The planar vehicle dynamics are invariant to $(x,y,\psi)$, whereas the arm's
inertia, gravity load, joint limits, and PD actuator torque are all functions of
$q$: the same driver command applied at different $(x,y,\psi)$ produces the same
change in body velocity, whereas the same joint command $q^{\mathrm{cmd}}$ at the
same $\dot q$ produces a different joint response at different $q$. An ablation with the
same architecture, data, and recipe confirms this: a variant that receives only
$\dot q$ and integrates $q$ from the predicted velocity raises the open-loop
end-effector drift from $0.3\%$ to $3.1\%$ err/dist at $0.5$\,s and from $1.2\%$
to $30\%$ at $2$\,s, and even a variant that receives $q$ but integrates it the
same way is $3$--$4\times$ worse than the deployed model, since $50$\,Hz velocity
samples miss the intra-step motion of the stiff PD joints.

\subsection{Model Training and Accuracy}
\label{subsec:tracked-training}

Both models instantiate the shared causal transformer of
Sec.~\ref{subsec:framework-architecture} over continuous state--action tokens at a
$50$\,Hz ($\Delta t = 0.02$\,s) sampling interval with a $16$-step
($\approx 0.32$\,s) context, predicting the reduced-state increment and advancing
$z_{t+1}=z_t+\Delta z_t$. Their state definitions and model capacities are
selected according to their respective tasks
(Table~\ref{tab:tracked-arm-config}): the tracked-vehicle model is a compact
$3$-layer / $4$-head / $96$-D transformer ($0.34$\,M parameters), while the arm
model is a larger $5$-layer / $8$-head / $256$-D transformer ($4.0$\,M
parameters). The arm configuration is the exact dynamics model inside which the
reaching policy is trained (Sec.~\ref{subsec:tracked-policies}), so every arm
number reported here characterizes that same deployed model.

\begin{table}[t]
  \centering
  \footnotesize
  \caption{Task-specific NRD-model configurations. Both instantiate the same
  causal next-state framework, while their state definitions and model capacities
  are selected according to their respective tasks. The arm column is the dynamics
  model used to train the reaching policy.}
  \label{tab:tracked-arm-config}
  \setlength{\tabcolsep}{10pt}
  \begin{tabular}{@{}l r r@{}}
    \toprule
    Item & Tracked vehicle & Arm \\
    \midrule
    Physical task            & planar goal reach & end-effector reach \\
    Reduced-state dim        & $3$ & $8$ \\
    Action dim               & $3$ & $4$ \\
    Sampling interval        & $0.02$\,s & $0.02$\,s \\
    Context length           & $16$ & $16$ \\
    Transformer layers       & $3$ & $5$ \\
    Attention heads          & $4$ & $8$ \\
    Embedding dim            & $96$ & $256$ \\
    Parameters               & $0.34$\,M & $4.0$\,M \\
    Training episodes        & $2{,}160$ & $15{,}000$ \\
    Training transitions     & $1.41$\,M & $0.76$\,M \\
    Training epochs          & $40$ & $80$ \\
    \bottomrule
  \end{tabular}
\end{table}

\begin{figure}[t]
  \centering
  \includegraphics[width=0.95\textwidth]{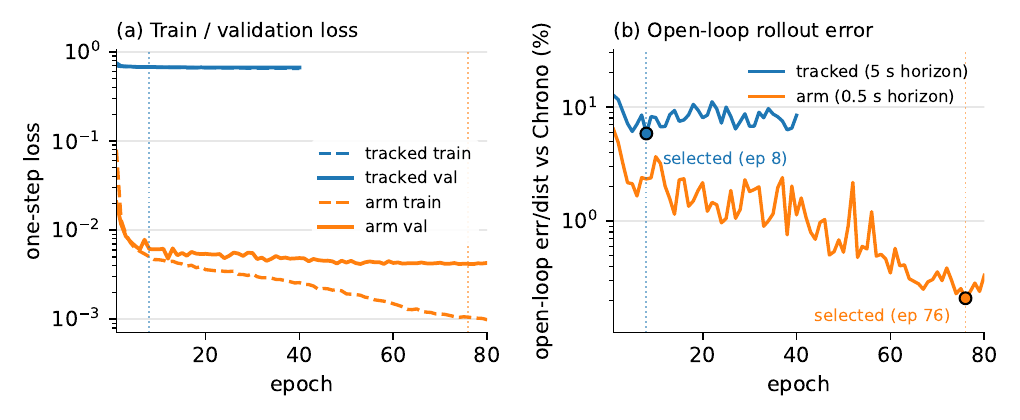}
  \caption{NRD-model training for both Case Study~II models (tracked vehicle,
  blue; arm, orange), the tracked vehicle over $40$ epochs and the arm over $80$.
  (a)~train (dashed) and validation (solid) one-step loss, log scale; (b)~open-loop
  rollout error against Chrono ground truth (err/dist, log scale)---the tracked
  vehicle scored on the $5$\,s horizon and the arm on the $0.5$\,s horizon. Each
  frozen model is selected by \emph{minimum open-loop rollout error} (marker;
  dotted line).}
  \label{fig:tracked-training}
\end{figure}

We train each model with the shared objective and causal transformer of
Sec.~\ref{sec:framework} using AdamW with a cosine-decayed learning rate---the
tracked-vehicle model over $40$ epochs and the arm model over $80$ epochs---and
freeze the checkpoint of lowest open-loop rollout error against Chrono rather
than the lowest one-step validation loss. One-step loss scores each transition in
isolation, whereas downstream control depends on the model staying accurate
across a rollout, so selecting on the multi-step open-loop error targets that
quantity directly. Figure~\ref{fig:tracked-training} reports both signals:
panel~(a) the train and validation one-step loss and panel~(b) the open-loop
rollout error (err/dist) used for selection, with the chosen epoch marked ($8$ for
the tracked vehicle, $76$ for the arm). The arm loss converges smoothly---its
validation plateaus near $4\times10^{-3}$ with a small train--validation gap and
its rollout error falls by an order of magnitude---while the tracked-vehicle one-step
loss is noise-limited and nearly flat, so its fidelity is judged from the
open-loop rollout error rather than the loss magnitude.

Unlike the HMMWV case, whose heterogeneous terrain dynamics motivated an
architecture sensitivity study, the tracked-vehicle and arm models operate on
substantially smaller reduced state spaces. We therefore use compact transformer
configurations selected from preliminary validation and do not perform a separate
architecture optimization. The purpose of Case Study~II is to evaluate whether
task-specific reduced abstractions support downstream control, rather than to
identify an optimal neural architecture.

Both models are judged by held-out one-step error and, more importantly, by
open-loop rollout error normalized by distance traveled (err/dist), the quantity
that governs downstream policy learning and transferring back to Chrono. Table~\ref{tab:tracked-arm-accuracy}
reports both. For the tracked vehicle, the model is rolled open-loop on recorded
driver commands over held-out episodes and the pose is integrated from the
predicted velocities; the $5$\,s rollout reaches $0.105$\,m position RMSE at
$5.85\%$ err/dist and $1.39^\circ$ yaw RMSE, so the integrated trajectory stays
close over a full maneuver. For the arm, the model is rolled open-loop from a
seeded context and the end-effector recovered by forward kinematics from the
predicted joints is compared to ground truth; end-effector drift stays near
$0.3\%$ err/dist out to $0.5$\,s and grows to only $\approx 1.2\%$ at $2$\,s. This
fidelity is sufficient to justify training policies inside the frozen models.

\begin{table}[t]
  \centering
  \footnotesize
  \caption{NRD-model accuracy. Vehicle one-step is the per-channel one-step
  RMSE of all three reduced states ($v_x$, $v_y$ in m/s; yaw rate $r$ in rad/s);
  vehicle open-loop is the $5$\,s pose-integrated position RMSE / err/dist. Arm one-step is the
  end-effector position RMSE (forward kinematics on the predicted joints); arm
  open-loop is the end-effector-drift err/dist at $2$\,s ($0.3\%$ at $0.5$\,s). All
  arm numbers are for the deployed $5$-layer model used to train the reaching
  policy (Table~\ref{tab:tracked-arm-config}).}
  \label{tab:tracked-arm-accuracy}
  \setlength{\tabcolsep}{8pt}
  \begin{tabular}{@{}l l l@{}}
    \toprule
    Model & One-step & Open-loop err/dist \\
    \midrule
    Tracked Vehicle NRD model & $v_x\,0.024$, $v_y\,0.013$\,m/s; $r\,0.0069$\,rad/s & $5.85\%$ / $0.105$\,m XY \\
    \midrule
    Arm NRD model  & $1.2$\,mm (ee)        & $1.2\%$ ee drift \\
    \bottomrule
  \end{tabular}
\end{table}

\subsection{Goal-Reaching Policies in the Frozen NRD Models}
\label{subsec:tracked-policies}

Both control tasks are goal-reaching problems, and both are solved the same way:
a policy is trained \emph{entirely inside} the corresponding frozen NRD model with
vectorized PPO. The two deliberately share
one implementation and actor--critic topology (Table~\ref{tab:tracked-arm-ppo})---both actor and critic use a $[256,128,64]$ ELU multilayer perceptron with empirical
observation normalization; \texttt{tanh}-squashed actions mapped into physical
bounds; a shaped per-step reward with a terminal success bonus; and clipped PPO
with an adaptive-KL learning-rate schedule, GAE, and empirical advantage
normalization. Reusing a single policy-learning framework across two very different
subsystems is the point of this case study: the tasks share that framework and the
actor--critic topology, while the state/action interfaces and task-specific training
settings differ.

The two tasks differ in three ways that follow directly from their abstractions.
First, the \emph{interface and goal geometry}: the vehicle acts through its
driver commands at $10$\,Hz and reaches a planar location, whose pose is
integrated \emph{outside} the learned state from the NRD model's predicted body
velocities; the arm acts through joint-command increments at $50$\,Hz and reaches
a three-dimensional end-effector target recovered by forward kinematics from the
NRD model's predicted joints, $p^{\mathrm{ee}} = \mathrm{FK}(q)$, so the reaching error
follows directly from the rollout. Second, \emph{tolerance and difficulty}: the vehicle must stop within
$0.75$\,m of a goal $20$--$40$\,m away, whereas the arm must place its
end-effector within $0.05$\,m while keeping collision- and joint-limit-safe. Third,
\emph{scale}: the arm is trained across $4{,}096$ parallel rollouts, the vehicle
across $2{,}048$. We detail each task in turn.

\begin{table}[t]
  \centering
  \footnotesize
  \caption{Task-specific PPO configurations using a common implementation and
  actor--critic topology, each trained
  inside its frozen NRD model; both use \texttt{tanh}-squashed actions, GAE
  ($\lambda=0.95$), a clip ratio of $0.2$, and empirical observation
  normalization. \textsuperscript{\ddag}\,the vehicle run is scheduled for $3{,}000$
  iterations but converges to $\approx100\%$ NRD-model success within the first
  $\sim150$ iterations; the iteration-$1{,}500$ checkpoint (long converged) is used
  for transfer.}
  \label{tab:tracked-arm-ppo}
  \setlength{\tabcolsep}{10pt}
  \begin{tabular}{@{}l r r@{}}
    \toprule
    Setting & Vehicle policy & Arm policy \\
    \midrule
    Parallel environments   & $2{,}048$ & $4{,}096$ \\
    Control rate            & $10$\,Hz & $50$\,Hz \\
    Observation / action dim & $11$\,/\,$3$ & $26$\,/\,$4$ \\
    Actor/critic hidden     & $[256,128,64]$ & $[256,128,64]$ \\
    Learning rate           & $3\!\times\!10^{-4}$ & $1\!\times\!10^{-4}$ \\
    Adaptive KL target      & $0.01$ & $0.005$ \\
    Discount $\gamma$       & $0.995$ & $0.99$ \\
    Rollout steps/env       & $32$ & $64$ \\
    Epochs / minibatches    & $5$\,/\,$4$ & $3$\,/\,$16$ \\
    Entropy coefficient     & $0.005$ & $0.001$ \\
    Success tolerance       & $0.75$\,m & $0.05$\,m \\
    PPO iterations          & $1{,}500$\textsuperscript{\ddag} & $1{,}500$ \\
    \bottomrule
  \end{tabular}
\end{table}

\subsubsection{Tracked-Vehicle Planar Goal Reaching}
\label{subsubsec:tracked-vehicle-policy}

The vehicle policy drives the platform from rest at the origin to a goal sampled at
radius $20$--$40$\,m and arbitrary bearing. At each control step it receives the
$11$-dimensional observation
\begin{equation}
  \label{eq:tracked-vehicle-obs}
  o^{\mathrm{veh}}_t =
  \big[\,\hat v_x,\hat v_y,\hat r,\ g_x^{b},g_y^{b},d,\
  \sin\theta_e,\cos\theta_e,\ \hat a_{t-1}\,\big] \in \R^{11},
\end{equation}
the standardized body velocities $(\hat v_x,\hat v_y,\hat r)$, the goal in the
body frame and its range $(g_x^{b},g_y^{b},d)$ scaled by $10$\,m, the sine and
cosine of the heading error $\theta_e$, and the normalized previous action. The
policy emits $a=[\delta,\tau,\beta]$ (steering, throttle, brake), each
\texttt{tanh}-squashed and mapped into their physical ranges, with throttle and braking capped at
$0.6$; one command is held for five dynamics steps, giving $10$\,Hz control over
the $50$\,Hz NRD model. The per-step reward is a progress term with small regularizing
penalties,
\begin{equation}
  \label{eq:tracked-vehicle-reward}
  \begin{aligned}
  r^{\mathrm{veh}}_t = {}& w_p\,\Delta d - w_\psi|\theta_e|
      - \lambda_a\lVert a_t-a_{t-1}\rVert^2
      - \lambda_{tb}\,\tau_t\beta_t \\
    &{} - \lambda_s\big(\max(|r|-\bar r,0)\big)^2
      - c + b\,\mathbf{1}[\,d<0.75\,],
  \end{aligned}
\end{equation}
where $\Delta d$ is the reduction in goal distance over the step. We use progress
weight $w_p{=}10$, heading $w_\psi{=}0.05$, action-rate $\lambda_a{=}0.01$,
throttle--brake $\lambda_{tb}{=}0.05$, a spin penalty $\lambda_s{=}0.5$ beyond
$\bar r{=}0.4$\,rad/s, a constant time penalty $c{=}0.1$ per step, and a terminal
success bonus $b{=}50$. The spin and throttle--brake terms suppress the degenerate
in-place-rotation and drive-against-brake behaviors that a pure progress reward
would otherwise admit. Training converges quickly: the mean episode reward rises
from negative values to a plateau within $\sim250$ iterations
(Fig.~\ref{fig:tracked-arm-rl-reward}(a)), and the episode success rate reaches
$\approx100\%$ within the first $\sim150$. At the iteration-$1500$ checkpoint used for transfer, every NRD-model
episode succeeds ($100\%$, mean final distance $0.62$\,m, no out-of-bounds
terminations); the run reaches this in $\approx10$\,min of wall-clock on a single
GPU.

\subsubsection{Arm End-Effector Reaching}
\label{subsubsec:arm-policy}

The arm policy moves the end-effector to a goal $g$ in the reachable workspace.
Goals are drawn in joint space, mapped through forward kinematics, and
safety-screened. At each control step the policy receives the $26$-dimensional
observation
\begin{equation}
  \label{eq:arm-obs}
  o^{\mathrm{arm}}_t =
  \big[\,q,\ \dot q,\ q^{\mathrm{cmd}},\
  g,\ p^{\mathrm{ee}},\ g-p^{\mathrm{ee}},\
  \hat\kappa,\ \hat a_{t-1}\,\big] \in \R^{26}.
\end{equation}
Here $q,\dot q\in\R^{4}$ are the joint positions and velocities of the reduced
state $z^{\mathrm{arm}}$ (Eq.~\ref{eq:tracked-arm-state}), standardized by the
training statistics. $q^{\mathrm{cmd}}\in\R^{4}$ is the current absolute joint
setpoint, i.e., the actuator command the environment carries from step to step.
$g,p^{\mathrm{ee}}\in\R^{3}$ are the goal and end-effector positions in the
arm-base frame, and $g-p^{\mathrm{ee}}$ is the reaching error. $\hat\kappa\in\R$ is a clipped clearance signal, a forward-kinematics
measure of proximity to collision (Appendix~\ref{app:arm-safety}).
$\hat a_{t-1}\in\R^{4}$ is the normalized previous action.

The policy emits a \texttt{tanh}-squashed joint-command increment
$a_t=\Delta q^{\mathrm{cmd}}_t\in[-\Delta q_{\max},\Delta q_{\max}]^{4}$ at
$50$\,Hz, which the environment accumulates into the absolute setpoint
$q^{\mathrm{cmd}}$ that the NRD model consumes (Eq.~\ref{eq:tracked-arm-state}).
Two checks guard this setpoint. First, it is clipped to the joint limits. Second,
a geometric safety shield (Appendix~\ref{app:arm-safety}) samples five
configurations along the straight joint-space path from the current setpoint to
the proposed one, evaluates forward kinematics at each, and tests for contact
with the ground, the vehicle, or another arm link; if any sample fails, the
increment is discarded (the setpoint is held) and the policy is penalized. The
policy thus learns joint-limit and collision avoidance \emph{inside} the NRD
rollout, from geometry alone, rather than discovering it later in Chrono.

The reward is an exponential reaching term with an action-rate regularizer and a
terminal bonus,
\begin{equation}
  \label{eq:arm-reward}
  r^{\mathrm{arm}}_t =
    \exp\!\big(-\lVert g-p^{\mathrm{ee}}\rVert/\sigma\big)
    - \lambda_a\lVert \hat a_t-\hat a_{t-1}\rVert^2
    + b\,\mathbf{1}[\,\lVert g-p^{\mathrm{ee}}\rVert<0.05\,].
\end{equation}
Here $\sigma{=}0.15$\,m is the reaching scale, $\lambda_a{=}0.02$ the action-rate
weight, and $b{=}150$ the success bonus, granted once the end-effector is within
the $0.05$\,m tolerance. The end-effector position is recovered from the NRD
model's predicted joints as $p^{\mathrm{ee}}=\mathrm{FK}(q)$, using the same forward
kinematics the safety shield already evaluates each step, so the reward reads
directly from the rollout at no extra cost.

\begin{figure}[t]
  \centering
  \includegraphics[width=0.95\textwidth]{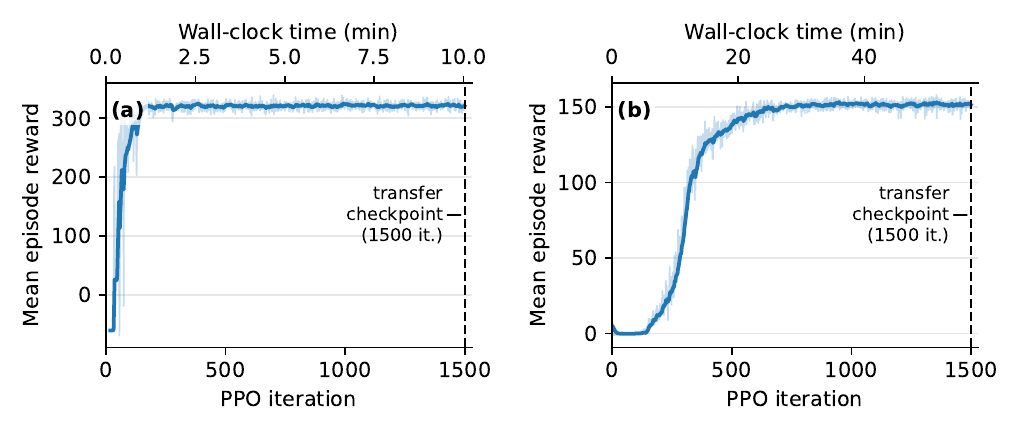}
  \caption{PPO convergence for the two goal-reaching policies, each trained inside
  its frozen NRD model: (a)~tracked-vehicle planar reaching and (b)~arm end-effector
  reaching. Each panel shows the mean episode reward versus PPO iteration; the top
  axis converts iteration to that run's wall-clock training time on a single GPU
  (the arm run is $\approx6\times$ longer). Faint traces are raw per-iteration
  values and solid curves are EMA-smoothed. The vehicle reward saturates almost
  immediately, whereas the tighter $0.05$\,m arm task ramps up more slowly to its
  plateau. The dashed line marks the checkpoint used for Chrono transfer.}
  \label{fig:tracked-arm-rl-reward}
\end{figure}

Reaching is the harder of the two tasks, and its convergence reflects that. The
mean episode reward ramps up over roughly $250$--$600$ iterations to a
$\approx150$ plateau (Fig.~\ref{fig:tracked-arm-rl-reward}(b)), and the episode
success rate settles at $\approx96.9\%$. Safety violations stay rare throughout:
at the iteration-$1500$ checkpoint the shield-flagged collision rate is well under
$1\%$ and joint limits are never hit, so the shield shapes behavior without
dominating it. The run trains across $4{,}096$ parallel rollouts in
$\approx57$\,min on a single GPU.

\subsection{Closed-Loop Chrono Transfer}
\label{subsec:tracked-transfer}

Chrono transfer is the decisive test: each policy, trained only inside its frozen
NRD model, is deployed in the full high-fidelity vehicle-plus-arm scene. The
policy is memoryless in both cases, so Chrono needs no dynamics context---it reads
the true system state, forms the same observation the policy trained on, and
applies the resulting command.

\paragraph{Tracked-vehicle transfer.} The vehicle policy is evaluated on a seeded
$100$-goal stress battery sampled from its training region (radius $20$--$40$\,m,
full angle $\pm\pi$), with one freshly built Chrono scene per goal. The policy reaches all $100$ goals
within the $0.75$\,m tolerance, though this is a coarse criterion: the policy stops
on entering the goal region rather than homing onto the goal, so it hugs the
tolerance radius (median closest approach $0.691$\,m, worst $0.748$\,m). Median
time-to-success is $20.2$\,s (worst $27.7$\,s) and no episode times out. The driven
paths are near-direct---median path efficiency $0.959$ (straight-line over driven
distance)---and goals behind the vehicle are serviced by clean forward-loop
U-turns rather than in-place spins, so the NRD-model policy transfers to the
full high-fidelity system across the goal distribution, if to this loose tolerance
(Appendix~\ref{app:stress}).

\paragraph{Arm transfer.} The arm policy is evaluated on a seeded $100$-goal
battery drawn from its trained joint-space workspace, one freshly built Chrono
scene per goal after a settle pre-roll. It reaches $97$ of $100$ goals within the
$0.05$\,m tolerance, with median reached end-effector error $4.2$\,cm (worst
$5.0$\,cm) and median convergence $0.9$\,s. All three misses are timeouts at deep
lower-workspace goals (target height down to $-4.4$\,m in the arm-base frame; closest
approach $6.4$, $9.3$, and $10.9$\,cm)---a region the training data under-samples,
not a safety failure: every accepted command stays collision- and joint-limit-safe
across all $100$ Chrono rollouts, with \emph{zero} contacts and \emph{zero}
joint-limit violations. The concentration of all three misses in the under-sampled
deep lower workspace is consistent with a data-coverage limitation; none is
associated with a contact or joint-limit violation, and the forward-kinematics
safety shield carries from training into the high-fidelity system
(Appendix~\ref{app:stress}).

\begin{table}[t]
  \centering
  \footnotesize
  \caption{Chrono transfer summary over the seeded $100$-goal stress batteries
  (Appendix~\ref{app:stress}). Vehicle: $100/100$ within $0.75$\,m, median closest
  approach $0.69$\,m (worst $0.748$\,m), median time-to-success $20.2$\,s, median
  path efficiency $0.959$. Arm: $97/100$ within $0.05$\,m (median reached error
  $4.2$\,cm), all three misses timeouts at under-sampled deep lower-workspace goals
  with zero contacts or joint-limit violations. Error scales differ
  because the two tasks use very different success tolerances.}
  \label{tab:tracked-arm-transfer}
  \setlength{\tabcolsep}{10pt}
  \begin{tabular}{@{}l r r r@{}}
    \toprule
    Task & Trials & Success & Median final err \\
    \midrule
    Vehicle goal reaching & $100$ & $100\%$ & $0.69$\,m \\
    Arm reaching       & $100$ & $97\%$  & $0.042$\,m \\
    \bottomrule
  \end{tabular}
\end{table}

\subsection{Case Study Takeaway}
\label{subsec:tracked-takeaway}

Case Study~II is not meant to establish that either task is an individually
difficult control benchmark. It shows that the proposed NRD methodology is reusable
across subsystems with different dominant dynamics and different control
objectives. The planar vehicle model retains only the quantities that propagate
vehicle pose, whereas the arm model retains only the joint positions and velocities on which
the joint response depends, takes the actuator command as its action, and recovers
the task-space end-effector by forward kinematics. In
both cases, a policy trained entirely inside the corresponding frozen NRD model
transfers to the high-fidelity Chrono system---reaching $100\%$ of $100$ seeded vehicle
goals at the coarse $0.75$\,m tolerance and $97\%$ of $100$ seeded arm goals at the
tight $0.05$\,m tolerance. Together with Case Study~I,
these results support the paper's claim that neural reduced dynamics should be
designed around the control task rather than around a fixed notion of system
state---the theme developed next in Sec.~\ref{sec:cross-case}.
\section{Analysis}
\label{sec:cross-case}

The two case studies approach the same question from different directions. Case
Study~I provides depth: an abstraction can fail when control-relevant
terramechanics variables or regime context are omitted. Case Study~II provides
breadth: the same workflow carries over when the appropriate abstraction changes
substantially across tasks and subsystems. Together they support a single account
of what a \emph{right} abstraction is---one that follows the task's dominant
physics, preserves the fidelity control requires, takes on context only where the
reduced dynamics are ambiguous, and earns its keep through throughput---and that
is judged, in the end, by whether the policies it trains transfer back into the
high-fidelity simulator. Table~\ref{tab:cross-summary} lays this pattern out
across the three tasks.

\begin{table*}[t]
  \centering
  \footnotesize
  \renewcommand{\arraystretch}{1.25}
  \caption{The same design pattern across three control tasks: each keeps a
  task-specific reduced state, treats commands and context as inputs, and recovers
  known kinematics analytically---yet every one yields a policy that transfers into
  high-fidelity Chrono.}
  \label{tab:cross-summary}
  \begin{tabularx}{\textwidth}{@{}L{1.7cm} Y Y Y Y Y@{}}
    \toprule
    Control task & Learned recurrent state & Action / context
      & Evaluated outside the NRD model & Why this abstraction is needed & Chrono evidence \\
    \midrule
    HMMWV tracking
      & Body velocities, attitude, and angular rates; per-wheel normal loads and speeds ($15$-D)
      & Steering, throttle, braking; rigid/CRM terrain code
      & Vehicle pose $(x,y,\psi)$; body frame error from reference trajectories
      & Exposes wheel slip, load transfer, and the domain-dependent tire--terrain response
      & Lower median and mean tracking error than both single-terrain specialist policies on flat, CRM, and zero-shot bumpy terrain \\
    \addlinespace[2pt]
    Tracked vehicle reaching
      & Planar velocities and yaw rate ($3$-D)
      & Steering, throttle, braking
      & Vehicle pose $(x,y,\psi)$; body frame error from goal point
      & Skid-steer chassis response in the plane is pose-invariant; pose is integrable
      & $100/100$ goals within $0.75$\,m \\
    \addlinespace[2pt]
    Arm reaching
      & Joint positions and velocities ($8$-D)
      & Joint target command $q^{\mathrm{cmd}}$
      & End-effector $p^{\mathrm{ee}}=\mathrm{FK}(q)$ and contact clearance
      & Joint response is dynamic and depends on $q$; task geometry is known analytically
      & $97/100$ goals within $0.05$\,m; zero contacts or joint-limit violations \\
    \bottomrule
  \end{tabularx}
\end{table*}

\subsection{Abstraction Follows the Task and Dominant Physics}
\label{subsec:cross-varies}

The three NRD models learn three different states. The terrain-aware HMMWV
carries a fifteen-dimensional state that couples body motion and attitude
($v_x,v_y,\phi,\theta$), the three body-frame angular rates, and an
eight-channel terramechanics block of per-wheel normal loads and wheel speeds
(Eq.~\eqref{eq:state-15d}). The tracked vehicle keeps only the three planar
quantities its goal-reaching controller acts through---the two body-frame
velocities and the yaw rate, $[v_x,v_y,r]$ (Eq.~\eqref{eq:tracked-vehicle-state}).
The arm keeps the eight variables that describe its configuration and how fast
it is changing, the joint positions and velocities $[q,\dot q]$
(Eq.~\eqref{eq:tracked-arm-state}).

The retained variables differ in kind, not only in count: each controller needs
different \emph{physical information}, and the physics that dominates each
control problem differs.
For the HMMWV the decisive and uncertain physics is the tire--terrain
interaction: whether the vehicle holds a reference across flat rigid ground,
bumpy rigid ground, and deformable CRM soil is strongly influenced by
terrain-dependent load transfer and wheel slip, so the reduced state should
expose the tire normal loads and wheel speeds. For the tracked vehicle the task is
planar goal reaching within a fixed operating envelope; the controller must
anticipate how steering, throttle, and braking produce longitudinal, lateral, and
yaw motion through the tracked chassis, and the three retained velocities capture
the response required to propagate planar pose. For the arm, reaching is a
configuration-space problem: the joints accelerate under actuator torque as they
are driven toward a commanded setpoint, so their future motion is set by where
the joints are and how fast they are moving---the joint positions and
velocities. Because the vehicle is held stationary in reach mode, its motion and
tire--terrain variables need not be represented in the arm NRD model.

The lesson is not that a smaller state is better, nor that there is a single
correct abstraction to be found. Each abstraction is a compact recurrent state
shown by rollout accuracy and closed-loop transfer to retain sufficient
control-relevant information for its task, and the dimension is a consequence of
that choice rather than the object of it. The
right abstraction is task-dependent because the dominant physics is
task-dependent.

\subsection{Reduction Preserves Task-Relevant Fidelity}
\label{subsec:cross-fidelity}

Reduction is as much a decision about what \emph{not} to learn as about what to
learn. In each case the model propagates only the coordinates on which the transition
depends and whose evolution is uncertain or contact-dependent, the part of the
dynamics that known relationships (such as FK or coordinate transformation) cannot
supply, and leaves the coordinates the transition does not depend on, together
with deterministic functions of the predicted state, to be reconstructed
analytically during rollout.

The pattern is consistent across the three systems. For both vehicles the global
pose is never predicted: the HMMWV's absolute position and heading, and the
tracked vehicle's planar $(x,y,\psi)$, are reconstructed by integrating the
predicted body velocities and yaw rate outside the learned state
(Sec.~\ref{subsec:hmmwv-state}, Sec.~\ref{subsec:tracked-abstractions}). For the
arm, the end-effector position is not a learned channel but is recovered by
forward kinematics on the predicted joints, $p^{\mathrm{ee}}=\mathrm{FK}(q)$; the
collision clearance the policy observes is computed geometrically by the same
kinematics (Appendix~\ref{app:arm-safety}); and the joint target command
$q^{\mathrm{cmd}}$, which a low-level PD controller drives the joint angles
toward, is supplied to the model as an input (Eq.~\eqref{eq:tracked-arm-state}).
The arm's joint positions could likewise be integrated from the predicted $\dot q$,
but the joint response depends on them, so they are retained in the state
(Sec.~\ref{subsec:tracked-abstractions}); what is reconstructed is decided by
whether the transition depends on a quantity, not by whether it is integrable. In
every case the learned model carries only the recurrent variables on which the
uncertain dynamics depend, and known kinematic and geometric relationships are
evaluated analytically rather than represented by additional learned channels.

The learned channels are a
projection of the same high-fidelity simulator---trained on its data and checked
against it in open- and closed-loop---so they retain fidelity where it is hardest
to obtain. The reconstructed quantities remain analytically consistent with the
predicted reduced state and require no additional learned output channel; their
accuracy relative to Chrono still depends on the accuracy of that predicted
state, and the collision-clearance calculation is an approximate geometric
screening measure. What reduction discards is not fidelity but
redundancy: state that either does not affect the control task, or that the
transition does not depend on and can be recovered without learning. The resulting model does not reproduce the
simulator's full state, and it does not need to; it preserves the fidelity the
control task requires.

\subsection{Context Is Required When Reduced Dynamics Are Ambiguous}
\label{subsec:cross-context}

A practical advantage of a learned NRD model over a fixed analytical one is
that its data-driven nature lets a single model absorb experience from
physically distinct domains: rather than deriving a separate model per regime, we
train one backbone on data pooled across regimes and let it represent their
shared structure and their differences together. Realizing that advantage rests
on two of the design choices this paper isolates---the right abstraction and the
right context. The abstraction (Sec.~\ref{subsec:cross-varies}) gives the regimes
a common state in which their shared vehicle dynamics are represented once; the
context tells the model, when that shared state is not enough, which regime it is
currently in.

Context becomes necessary when the reduced dynamics are \emph{ambiguous}: when
the same reduced state and action evolve to different next states under different
physical regimes. In the HMMWV case a given body-and-terramechanics state under a
given driver command advances differently on rigid ground than on CRM soil,
because the two regimes generate different slip and sinkage
(Sec.~\ref{subsec:hmmwv-conditioning}). Without an explicit label the model must
infer the regime implicitly from history, which can blur the learned transition
and accumulate error over rollout. A small
two-class one-hot code---rigid $[1,0]$, CRM $[0,1]$, appended to each input
token---resolves the ambiguity, letting one backbone reuse the common
vehicle-dynamics representation while specializing the rigid-versus-CRM
transition. The context ablation makes the effect concrete: removing the code
barely changes one-step loss but more than doubles open-loop error on rigid
terrain, from $3.73\%$ to $8.26\%$, while leaving CRM error approximately
unchanged (Appendix~\ref{app:feature-ablation}). The history alone therefore does
not reliably preserve the rigid/CRM distinction over long rollouts.

\subsection{Throughput Makes Policy Learning Practical}
\label{subsec:cross-throughput}

Reduction earns its keep not only through accuracy but through speed, because the
NRD model replaces Chrono as the environment the policy trains in. A PPO run
consumes a large volume of environment transitions---thousands of parallel
rollouts advanced over hundreds to thousands of iterations---and the rate at
which those transitions are produced sets the wall-clock cost of training. The
NRD models run orders of magnitude faster than the simulator they stand in
for: batched on a single GPU the conditioned HMMWV model advances at
$\approx 1{,}750\times$ real time, the tracked vehicle at $\approx 16{,}300\times$,
and the arm at $\approx 2{,}320\times$ (Appendix~\ref{app:sim-speed}). Under the
reported aggregate batched-GPU versus single-process Chrono configurations, the
measured simulated-time throughput ratios reach about $11{,}500\times$ for CRM
soil, $58{,}400\times$ for the tracked vehicle, and $8{,}300\times$ for the arm; the
tracked-vehicle ratio is asymmetric, since Chrono simulates the full
vehicle-plus-arm scene while each surrogate models a single operating mode. That
margin turns policy
optimization from impractical into routine: the tracked vehicle driving policy reaches its
plateau in $\approx 10$\,min and the arm policy trains in $\approx 57$\,min on a
single GPU, each converging within a few hundred PPO iterations
(Sec.~\ref{subsec:tracked-policies}).

A single converged run, though, understates what throughput buys. The
time-consuming part of applying RL is not that run but the iterative design
around it---shaping the reward and tuning the optimizer until the policy behaves
as intended, each adjustment validated only by training again. Both policies
needed such work: spin and drive-against-brake penalties for the vehicle, and reward
shaping and safety-interface design for the arm---the terminal bonus, action
regularization, clearance observation, and command shield
(Sec.~\ref{subsec:tracked-policies}). The effective cost of a policy is therefore
one run times the number of design iterations, so a run that finishes in minutes
rather than the orders-of-magnitude greater wall-clock cost implied by the
measured single-process Chrono throughput keeps that loop interactive---more
reward and hyperparameter iterations per day, which is what ultimately yields a
working policy.

\subsection{High-Fidelity Validation Is the Final Test}
\label{subsec:cross-validation}

An NRD model is validated at three levels of increasing stringency. One-step
prediction is the most basic: given a ground-truth history, it checks whether the
model has captured the single-step input-to-output relation, and so tests the
soundness and capacity of the learned map. It is necessary but limited, because a
low one-step error does not guarantee that repeated application stays on
trajectory. Open-loop rollout is the more informative metric: feeding the model
its own predictions over a horizon measures how faithfully it reproduces Chrono's
behavior as a dynamical system, which is the fidelity that governs downstream
control. In both case studies the NRD model used to train RL is therefore chosen by
minimum open-loop rollout error rather than by one-step loss
(Sec.~\ref{subsec:hmmwv-conditioning}, Sec.~\ref{subsec:tracked-training}).
Closed-loop transfer into Chrono is the final level and the exam for the whole
pipeline: the trained policy parameters are transferred unchanged, while Chrono
reproduces the same observation construction, action processing, and safety
interface used during policy training inside the NRD model, testing the NRD model and the policy
together.

The third level is decisive, because the first two do not by themselves establish
that the abstraction is useful for control. A policy can earn high reward inside
an approximate model by exploiting its errors, so success in the NRD model
environment is necessary but not sufficient. What settles the question is Chrono:
the tracked vehicle reaches $100$ of $100$ seeded goals within $0.75$\,m and the arm
$97$ of $100$ within $0.05$\,m with zero contacts and joint-limit violations
(Sec.~\ref{subsec:tracked-transfer}), while the terrain-aware HMMWV policy
attains low tracking error on flat, CRM, and zero-shot bumpy
terrain---matching or improving on each single-terrain specialist even on its own
training terrain (Appendix~\ref{app:policy-transfer}). In every case the reduced
model is judged not by its own reward but by whether the policies it trains
remain effective in the full simulator, and that transfer is the ultimate
evidence that the abstraction was the right one.
\section{Limitations and Open Problems}
\label{sec:limitations}

The present work is validated entirely in simulation. Returning the learned
policies to Chrono tests whether each NRD model retains sufficient task-relevant
dynamics for control, but this remains sim-to-sim transfer and does not establish
performance on physical systems. The NRD models are trained on and evaluated
against a single high-fidelity reference, so whatever bias Chrono's tire,
terramechanics, and actuator models carry is inherited by the surrogate rather
than exposed by it. Both the NRD models and the policies also operate on compact
simulator states rather than sensory observations: the reduced state is read
directly from the simulator through the projection $P$
(Sec.~\ref{subsec:framework-problem}), and the derived quantities the policy
observes---vehicle pose, end-effector position, and clearance---are computed from
that privileged state, thereby bypassing perception and state-estimation errors.
Extending the framework to vision-based dynamics models and policies, in the
manner of latent world models learned from
observations~\cite{hafner2019planet,hafner2020dreamer}, followed by hardware
validation, is an important next step.

The construction of each reduced state also remains partly heuristic. The design
rule of Sec.~\ref{subsec:framework-architecture} states what the reduced state
should retain, but applying it---selecting recurrent variables, analytically
reconstructed quantities, context inputs, and operating ranges---requires physical
insight and task-specific experimentation. Although this process produces
interpretable abstractions, it requires substantial manual effort for each new
system. In future work we will introduce an automated or agentic abstraction-design
procedure that proposes candidate state projections, reconstruction mappings, and
context variables, and evaluates them jointly using rollout accuracy, policy
performance, transfer fidelity, and computational cost. The validation hierarchy
used here (Sec.~\ref{subsec:cross-validation}) already supplies the selection
signal such a search would optimize.

Finally, the framework was not evaluated on explicit contact-mode transitions.
Although the vehicle cases include aggregate tire--terrain interaction, the arm
model is trained only on free-space motion and its controller avoids contact
through the safety shield (Appendix~\ref{app:arm-safety}); grasp acquisition and
release, impacts, and object--object collisions therefore remain outside the
present scope. These events introduce discontinuous and potentially multimodal
dynamics that are difficult to represent reliably with a single continuous learned
transition model. Extending the method to contact-rich manipulation will likely
require event- or contact-conditioned neural dynamics, in which a contact mode is
predicted and the transition is conditioned on
it~\cite{jing2026contactaware,xu2025nerd}, combined with reduced deformable
simulation approaches such as Simplicits and
FreeForm~\cite{modi2024simplicits,xiang2026freeform} for the deformable and
contact-mediated physics that a compact rigid-state abstraction cannot represent.

\section{Conclusion}
\label{sec:conclusion}

This paper presented a framework for distilling high-fidelity robot simulations
into task-specific neural reduced dynamics models that serve as massively parallel
environments for policy learning. The central result is that useful reduction is
determined by control relevance rather than dimensionality alone. Each abstraction
retains the recurrent physical variables and context needed to propagate task
behavior, reconstructs known kinematic quantities analytically outside the
network, and is evaluated through sustained open-loop rollouts and, ultimately,
closed-loop transfer back to Chrono.

Across the HMMWV, tracked-vehicle, and arm tasks, three distinct abstractions
supported policy optimization entirely within frozen, vectorized NRD models. The
HMMWV generalist achieved lower median and mean tracking error than both
single-terrain specialists on every evaluated terrain, including the zero-shot
bumpy case. When transferred to Chrono, the tracked-vehicle and arm policies reached
$100$ of $100$ goals within $0.75$\,m and $97$ of $100$ within $0.05$\,m,
respectively, with no arm contacts or joint-limit violations. In both case studies
the NRD model advanced its task roughly four orders of magnitude faster in
simulated time than the high-fidelity scene it replaced, under the reported
batched-GPU surrogate and single-process Chrono configurations: $\approx
11{,}500\times$ for CRM soil in Case Study~I, and $\approx 8{,}300$--$58{,}400\times$
across the two control modes of Case Study~II. Together, these results show that, within the tested simulator
regimes, task-designed NRD models can retain the dynamics required for control while
moving high-volume policy optimization out of the expensive high-fidelity
simulator. Chrono remains the high-fidelity reference and final validation
authority.

\printcredits

\section*{Acknowledgment}
This work was in part supported by NSF grant CMMI-2153855.

\appendix
\section{Simulation Throughput and Real-Time Performance}
\label{app:sim-speed}

The neural reduced dynamics model is motivated by the wall-clock cost of the
high-fidelity Chrono simulator, which grows sharply with contact-model
complexity. We probed the stepping throughput of each Chrono scene used in this
paper with the same collector/evaluation scene builders that generate the data
(measured on a single Intel~14900KF, one process, throttle $0.4$), and
separately measured each case study's NRD model throughput as a batched GPU
forward pass during its own PPO training run. The cross-case comparable metric
is the real-time factor (RTF), i.e., simulated seconds advanced per wall-clock
second ($\mathrm{RTF}=1$ is real time); Cost is normalized to each case's
NRD model ($1\times$), so it reads directly as how many times more wall-clock
time high-fidelity simulation costs than the surrogate that replaces it.
Results are split by case study: Table~\ref{tab:sim-speed-hmmwv} (Case
Study~I) and Table~\ref{tab:sim-speed-tracked} (Case Study~II).

\subsection{Case Study I: Terrain-Aware HMMWV}
\label{app:sim-speed-hmmwv}

\begin{table}[t]
  \centering
  \footnotesize

  \caption{Case Study I (terrain-aware HMMWV) throughput. Chrono rows share
  the fixed \texttt{HMMWV\_Full} configuration and differ only in the
  tire--terrain contact model. Cost is normalized to the NRD model row
  ($1\times$). \textsuperscript{\ddag}\,The NRD model row is an aggregate
  batched-GPU figure, not a single-stream process like the three Chrono rows
  above: Steps/s is dynamics forward passes/s summed across the $2{,}048$
  parallel rollouts used during PPO training
  (Sec.~\ref{subsec:hmmwv-policy-transfer}), read from the rsl-rl training
  logs (\texttt{Perf/total\_fps}, consistently $\approx 35{,}000$
  policy-control steps/s across all three $8$-layer policies, $\times$
  action\_repeat $=5$); RTF is the corresponding aggregate simulated-seconds
  advanced per wall-clock second.}
  \label{tab:sim-speed-hmmwv}
  \setlength{\tabcolsep}{9pt}
  \begin{tabular}{@{}l r r r r@{}}
    \toprule
    Configuration & $\Delta t$ & Steps/s & RTF & Cost \\
    \midrule
    HMMWV --- rigid flat  & $2.0$\,ms & $3{,}657$   & $7.31\times$   & $239\times$ \\
    HMMWV --- bumpy rigid & $2.0$\,ms & $3{,}324$   & $6.65\times$   & $263\times$ \\
    HMMWV --- CRM soil    & $0.5$\,ms & $304$       & $0.152\times$  & $11{,}500\times$ \\
    \midrule
    HMMWV --- NRD model ($8$-layer)\textsuperscript{\ddag} & $10$\,ms & $175{,}000$ & $1{,}750\times$ & $1\times$ \\
    \bottomrule
  \end{tabular}
\end{table}

Flat rigid HMMWV runs at $7.31\times$ real time; a bumpy heightmap costs only
$\sim$$9\%$ more (RTF $\to 6.65\times$), isolating the heightmap-collision
overhead from an otherwise identical vehicle and step size. CRM is the
dominant Chrono cost at $0.152\times$ RTF ($\sim$$14.1$\,M SPH particles,
$\sim$$10.8$\,M boundary markers, and a $4\times$ smaller step, not the
vehicle model), so a $15$\,s CRM episode takes $\sim$$100$\,s of wall-clock
time. The $8$-layer NRD model sustains $\approx 175{,}000$ dynamics passes/s aggregate
($\approx 1{,}750\times$ RTF) independent of terrain
(Sec.~\ref{subsec:hmmwv-conditioning}); normalized to it, even flat-rigid
Chrono costs $239\times$ more wall-clock time per simulated second, and CRM
costs $11{,}500\times$ more.

\subsection{Case Study II: Tracked Vehicle + Arm}
\label{app:sim-speed-tracked}

\begin{table}[t]
  \centering
  \footnotesize

  \caption{Case Study II (tracked vehicle $+$ arm) throughput, measured the
  same way as Table~\ref{tab:sim-speed-hmmwv}. The single Chrono
  vehicle-plus-arm scene serves both control modes; each NRD model row is its own
  $1\times$ reference, and the Chrono Cost cell lists the wall-clock ratio
  over the drive-mode\,/\,reach-mode NRD model respectively.
  \textsuperscript{\ddag}\,as in Table~\ref{tab:sim-speed-hmmwv}, both NRD model
  rows are aggregate batched-GPU figures (not single-stream like the Chrono
  row): Steps/s is dynamics forward passes/s during PPO, read from the rsl-rl
  logs as \texttt{Perf/total\_fps} $\times$ action\_repeat. Drive mode uses
  \texttt{tracked\_transformer\_v1} ($2{,}048$ rollouts,
  \texttt{Perf/total\_fps} $\approx 163{,}000$ $\times$ action\_repeat $5$);
  reach mode uses the larger \texttt{arm\_transformer\_8d\_v1} ($4{,}096$
  rollouts, \texttt{Perf/total\_fps} $\approx 116{,}000$, action\_repeat $1$).
  The Chrono scene simulates the full coupled vehicle-plus-arm system, whereas
  each NRD model represents only the subsystem its mode controls.}
  \label{tab:sim-speed-tracked}
  \setlength{\tabcolsep}{9pt}
  \begin{tabular}{@{}l r r r r@{}}
    \toprule
    Configuration & $\Delta t$ & Steps/s & RTF & Cost \\
    \midrule
    Tracked vehicle $+$ arm --- rigid & $0.5$\,ms & $558$ & $0.279\times$ & $58{,}400\times$\,/\,$8{,}300\times$ \\
    \midrule
    Tracked vehicle --- NRD model\textsuperscript{\ddag} & $20$\,ms & $815{,}000$ & $16{,}300\times$ & $1\times$ \\
    Arm --- NRD model\textsuperscript{\ddag}          & $20$\,ms & $116{,}000$  & $2{,}320\times$  & $1\times$ \\
    \bottomrule
  \end{tabular}
\end{table}

The tracked drive-mode NRD model is smaller than the HMMWV backbone ($3$ layers,
$96$-D embedding, $20$\,ms step) but still batched across $2{,}048$ rollouts,
sustaining $\approx 815{,}000$ dynamics passes/s ($\approx 16{,}300\times$
RTF); the larger reach-mode arm NRD model ($5$ layers, $256$-D, batched across
$4{,}096$ rollouts) sustains $\approx 116{,}000$ passes/s ($\approx
2{,}320\times$ RTF). The single-process Chrono tracked-vehicle-plus-arm
simulation, at $0.279\times$ RTF (reflecting the small integration step and
stiff solver that single-pin track contact requires), costs $\approx
58{,}400\times$ more wall-clock time per simulated second than the drive-mode
surrogate and $\approx 8{,}300\times$ more than the reach-mode surrogate---the
former is the largest gap of either case study, though the comparison is not
fully symmetric: the Chrono scene simulates the full coupled vehicle-plus-arm
system, whereas each NRD model represents only the subsystem its mode
controls.
\section{Per-Terrain Policy-Transfer Statistics}
\label{app:policy-transfer}

Table~\ref{tab:policy-transfer} reports the full closed-loop Chrono tracking
statistics summarized by Fig.~\ref{fig:policy-transfer-bars}: for each terrain and
policy, the median, mean, and interquartile range (IQR) of the XY RMSE over the $20$
held-out references. All three policies are trained on the $8$-layer dynamics
backbone (Sec.~\ref{subsec:hmmwv-conditioning}, Appendix~\ref{app:ablation-ofat})
and evaluated at their $1000$-iteration checkpoint under the same action clipping, and every cell completes
$20/20$ rollouts with no early terminations.

\begin{table}[t]
  \centering
  \footnotesize

  \caption{Closed-loop Chrono tracking statistics (XY RMSE, meters) for the
  mixture generalist and the rigid-only / CRM-only specialists,
  on held-out rigid flat, CRM, and zero-shot bumpy references; these are the numbers
  summarized by Fig.~\ref{fig:policy-transfer-bars}. Median, mean, and IQR are over
  $20$ references at the $1000$-iteration checkpoint, all $20/20$ successful. Bold
  marks the best median and mean in each terrain block; on this $8$-layer backbone
  the mixture generalist takes every bold cell, i.e.\ it achieves the lowest mean and
  median XY RMSE on all three evaluated terrains.
  \textsuperscript{\dag}\,bumpy is a strict zero-shot test: rigid code, no bumpy data
  in training or selection.}
  \label{tab:policy-transfer}
  \begin{tabular*}{0.55\textwidth}{@{\extracolsep{\fill}}llrrr@{}}
    \toprule
    Terrain & Policy & Median & Mean & IQR \\
    \midrule
    \multirow{3}{*}{Rigid flat}
      & Mixture generalist & \textbf{0.157} & \textbf{0.184} & 0.099--0.236 \\
      & Rigid-only         & 0.174          & 0.219          & 0.163--0.277 \\
      & CRM-only           & 0.232          & 0.259          & 0.189--0.328 \\
    \cmidrule(lr){1-5}
    \multirow{3}{*}{CRM}
      & Mixture generalist & \textbf{0.180} & \textbf{0.249} & 0.134--0.321 \\
      & Rigid-only         & 0.854          & 1.000          & 0.365--1.447 \\
      & CRM-only           & 0.231          & 0.361          & 0.174--0.521 \\
    \cmidrule(lr){1-5}
    \multirow{3}{*}{Rigid bumpy\textsuperscript{\dag}}
      & Mixture generalist & \textbf{0.149} & \textbf{0.229} & 0.099--0.247 \\
      & Rigid-only         & 0.187          & 0.238          & 0.159--0.290 \\
      & CRM-only           & 0.213          & 0.418          & 0.166--0.416 \\
    \bottomrule
  \end{tabular*}
\end{table}

\begin{figure*}[!t]
  \centering
  \includegraphics[width=\textwidth]{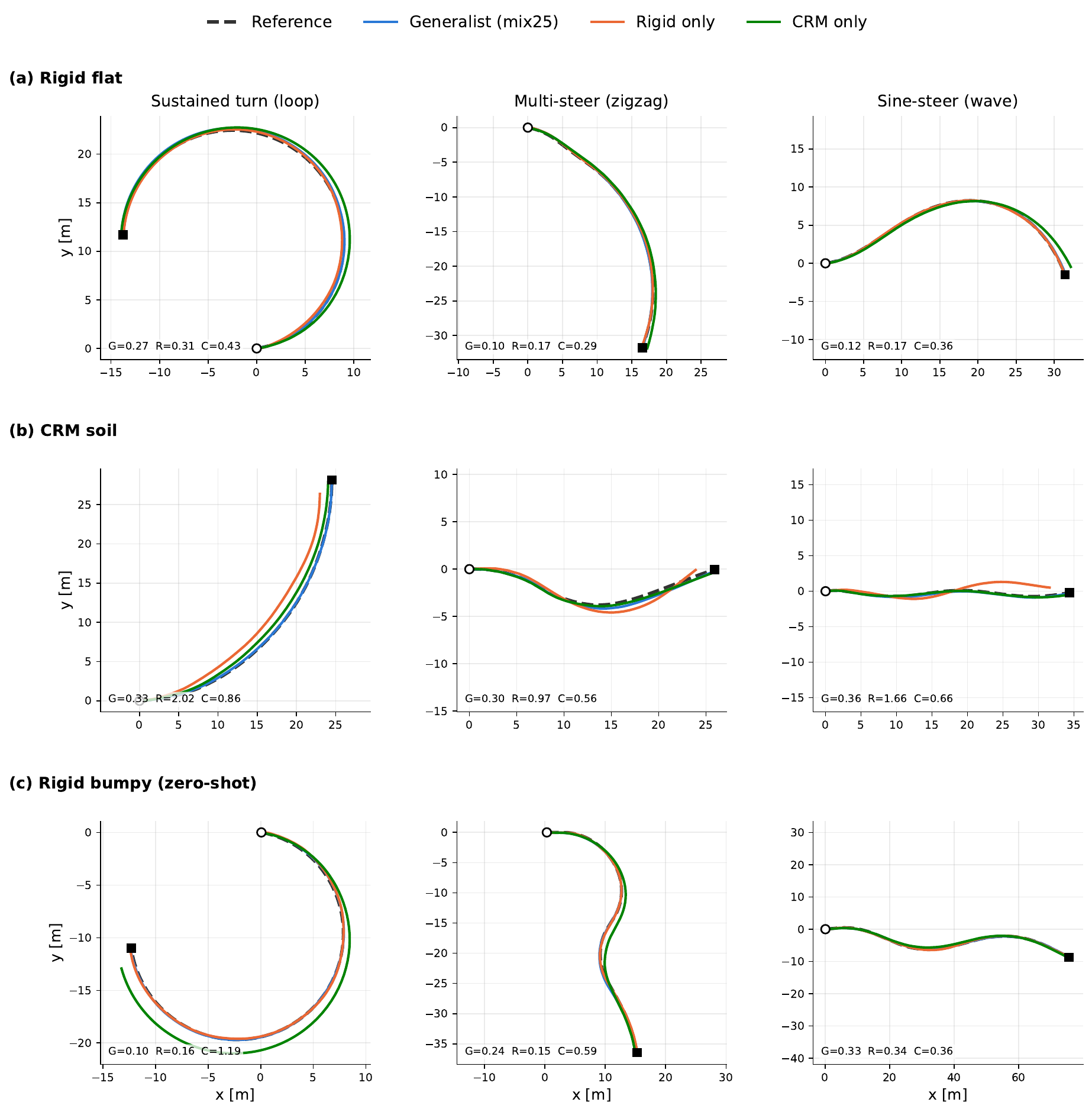}
  \caption{Closed-loop XY paths for three held-out maneuvers (columns) on each
  terrain (rows): (a) rigid flat, (b) CRM soil, (c) rigid bumpy (zero-shot).
  Each panel is annotated with the episode's XY RMSE (m) for the generalist
  (G), rigid-only (R), and CRM-only (C) policies. The rigid-only specialist
  (never trained on CRM) visibly overshoots every turn in row (b), consistent
  with its $1.000$\,m mean RMSE on CRM in Table~\ref{tab:policy-transfer}; the
  CRM-only specialist is the outlier in row (c) despite bumpy sharing its rigid
  contact code. Full $20$-reference statistics for every maneuver are in
  Table~\ref{tab:policy-transfer}.}
  \label{fig:policy-traj}
\end{figure*}

Figure~\ref{fig:policy-traj} shows the closed-loop Chrono XY paths behind these
statistics for three representative held-out maneuvers per terrain (sustained
turn, multi-steer, sine-steer; $9$ evaluations total), for the reference and all
three policies.

\section{Dynamics-Architecture Sweep}
\label{app:ablation-ofat}

The $8$-layer dynamics backbone used throughout this paper
(Sec.~\ref{subsec:hmmwv-conditioning}) is the result of a one-factor-at-a-time
(OFAT) architecture sweep over four axes: depth
($n_{\mathrm{layer}}\in\{2,4,6,8,12\}$), width ($n_{\mathrm{embd}}\in
\{128,192,256,384,512\}$, with the attention head count co-scaled so that
$\mathrm{head\_dim}=n_{\mathrm{embd}}/n_{\mathrm{head}}$ stays fixed at $32$),
attention-head width ($\mathrm{head\_dim}\in\{16,32,64\}$ at fixed
$n_{\mathrm{embd}}=256$), and train-time context
($\mathrm{block\_size}\in\{32,64,128,256\}$). Each
arm varies exactly one axis from a shared $6$-layer, $256$-embedding,
$\mathrm{head\_dim}$-$32$, $128$-context configuration (Ablation model $10$ in
Table~\ref{tab:ofat-sweep}),
holding the training recipe of Sec.~\ref{subsec:hmmwv-conditioning} fixed
otherwise; that shared configuration is trained once rather than once per axis, so
the four axes contribute $4$ (depth) $+\,4$ (width) $+\,2$ (heads) $+\,3$
(context) new configurations to the $1$ shared one, $14$ in total. Each is trained
for the full $80$ epochs and ranked by the same domain-balanced selection score
$S=\tfrac{1}{2}E_{\mathrm{rigid}}+\tfrac{1}{2}E_{\mathrm{CRM}}$ of
Eq.~\eqref{eq:worst-domain} used to pick every model's checkpoint.

\begin{table*}[t]
  \centering
  \footnotesize
  \renewcommand{\arraystretch}{1.15}
  \caption{All $14$ configurations in the dynamics-architecture sweep, sorted by
  selection score $S$ (Eq.~\ref{eq:worst-domain}; lower is better), each at its own
  best epoch. Each model varies exactly one of depth ($n_{\mathrm{layer}}$), width
  ($n_{\mathrm{embd}}$, with the attention head count co-scaled so
  $\mathrm{head\_dim}=n_{\mathrm{embd}}/n_{\mathrm{head}}$ stays fixed at $32$),
  attention-head width (head\_dim, at fixed $n_{\mathrm{embd}}=256$---the axis
  that actually changes head\_dim), or train-time context (ctx) from the shared
  $6$-layer/$256$-embedding/head\_dim-$32$/$128$-context configuration (Ablation
  model 10). \emph{One-step loss} is the channel-reweighted Huber validation loss as training
  (Eq.~\ref{eq:cotraining-loss}); \emph{rollout err/dist} is the $10$\,s open-loop
  position error normalized by distance traveled. The top row (bold), Ablation
  model 1, is the $8$-layer configuration used throughout this paper.}
  \label{tab:ofat-sweep}
  \begin{tabular*}{\textwidth}{@{\extracolsep{\fill}} l r r r r r r r r r @{}}
    \toprule
    Model & $n_{\mathrm{layer}}$ & $n_{\mathrm{embd}}$ & head\_dim & ctx
      & \multicolumn{2}{c}{One-step loss} & \multicolumn{2}{c}{10\,s rollout err/dist} & $S$ \\
    \cmidrule(lr){6-7}\cmidrule(lr){8-9}
    & & & & & Flat & CRM & Flat & CRM & \\
    \midrule
    \textbf{Ablation model 1 (used)} & \textbf{8} & \textbf{256} & \textbf{32} & \textbf{128}
      & \textbf{0.0025} & \textbf{0.094} & \textbf{3.7\%} & \textbf{5.4\%} & \textbf{4.6\%} \\
    Ablation model 2  & 6  & 256 & 32 & 256 & 0.0021 & 0.114 & 6.1\% & 3.7\% & 4.9\% \\
    Ablation model 3  & 6  & 512 & 32 & 128 & 0.0020 & 0.103 & 4.6\% & 5.3\% & 4.9\% \\
    Ablation model 4  & 6  & 384 & 32 & 128 & 0.0030 & 0.095 & 5.6\% & 4.9\% & 5.2\% \\
    Ablation model 5  & 6  & 256 & 16 & 128 & 0.0021 & 0.097 & 6.7\% & 3.9\% & 5.3\% \\
    Ablation model 6  & 12 & 256 & 32 & 128 & 0.0034 & 0.091 & 6.4\% & 4.8\% & 5.6\% \\
    Ablation model 7  & 6  & 192 & 32 & 128 & 0.0024 & 0.087 & 5.8\% & 6.1\% & 5.9\% \\
    Ablation model 8  & 2  & 256 & 32 & 128 & 0.0040 & 0.097 & 9.1\% & 4.4\% & 6.8\% \\
    Ablation model 9  & 6  & 256 & 64 & 128 & 0.0021 & 0.095 & 7.5\% & 6.1\% & 6.8\% \\
    Ablation model 10 & 6  & 256 & 32 & 128 & 0.0022 & 0.095 & 9.1\% & 5.8\% & 7.4\% \\
    Ablation model 11 & 6  & 256 & 32 & 64  & 0.0023 & 0.090 & 6.9\% & 8.3\% & 7.6\% \\
    Ablation model 12 & 4  & 256 & 32 & 128 & 0.0034 & 0.091 & 8.7\% & 7.0\% & 7.8\% \\
    Ablation model 13 & 6  & 256 & 32 & 32  & 0.0035 & 0.101 & 9.4\% & 6.9\% & 8.2\% \\
    Ablation model 14 & 6  & 128 & 32 & 128 & 0.0037 & 0.092 & 8.8\% & 7.7\% & 8.3\% \\
    \bottomrule
  \end{tabular*}
\end{table*}
\section{Training-Data Scaling}
\label{app:data-scaling}

Holding the $8$-layer backbone (Appendix~\ref{app:ablation-ofat}) and the rest of
the training recipe (Sec.~\ref{subsec:hmmwv-conditioning}) fixed, we retrain it on
nested seeded subsets of $20\%$, $40\%$, $60\%$, and $80\%$ of the training
episodes, applied to both the flat and CRM sources so the $75/25$ mix is
preserved; compute is held fixed at $80$ epochs $\times$ $2000$ mini-batches by
sampling with replacement over the smaller pool, so unique-trajectory count is the
only variable. The $100\%$ point is the same run used throughout the paper
(Ablation model 1 of Table~\ref{tab:ofat-sweep}).

\begin{table*}[t]
  \centering
  \footnotesize
  \renewcommand{\arraystretch}{1.15}
  \caption{Data-quantity ablation at the $8$-layer backbone: selection score $S$
  (Eq.~\ref{eq:worst-domain}) and its one-step-loss/rollout-err components at each
  run's own best epoch, as a function of the fraction of training episodes used.
  \textsuperscript{\dag}\,$80\%$ attains a marginally lower single-seed $S$ than
  the full-data point; see discussion below.}
  \label{tab:data-scaling}
  \begin{tabular*}{\textwidth}{@{\extracolsep{\fill}} l r r r r r @{}}
    \toprule
    & \multicolumn{2}{c}{One-step loss} & \multicolumn{2}{c}{10\,s rollout err/dist} & \\
    \cmidrule(lr){2-3}\cmidrule(lr){4-5}
    Data & Flat & CRM & Flat & CRM & $S$ \\
    \midrule
    20\%  & 0.0023 & 0.138 & 4.2\% & 9.6\% & 6.9\% \\
    40\%  & 0.0022 & 0.124 & 8.1\% & 4.8\% & 6.5\% \\
    60\%  & 0.0021 & 0.114 & 7.0\% & 5.7\% & 6.3\% \\
    80\%\textsuperscript{\dag}  & 0.0020 & 0.105 & 3.9\% & 4.6\% & 4.3\% \\
    \textbf{100\% (used)} & \textbf{0.0025} & \textbf{0.094} & \textbf{3.7\%} & \textbf{5.4\%} & \textbf{4.6\%} \\
    \bottomrule
  \end{tabular*}
\end{table*}

$S$ falls from $6.9\%$ to $4.3\%$ over the $20$--$80\%$ range---more training data
consistently buys rollout accuracy. The $100\%$ point does not continue that
trend in this single training seed ($S=4.6\%$, marginally above the $80\%$
point), which we read as seed noise rather than a genuine data-quantity
regression, since nothing else changes between the two runs and every other
step of the $20$--$80\%$ curve is monotonic. We use the full dataset as the
baseline reported throughout the paper: the scaling trend gives no reason to
expect held-back data to hurt, and $100\%$ removes the arbitrariness of choosing
a fixed held-out fraction of otherwise-usable training trajectories.
\section{Reduced-State and Context Ablation}
\label{app:feature-ablation}

Sec.~\ref{subsec:hmmwv-state} and Sec.~\ref{subsec:hmmwv-conditioning} motivate the
15-D reduced state and the one-hot terrain code from first principles---the four
tire normal forces and four wheel speeds are argued to be what makes slip and load
transfer observable, and the terrain key is argued to be what lets one shared
backbone disambiguate regimes. Here we test both claims directly by removing each
block from the $8$-layer backbone (Appendix~\ref{app:ablation-ofat}) and retraining,
holding the rest of the recipe fixed ($8$ layers, $8$ attention heads, $256$-D
embedding, $128$-step context, $75/25$ flat/CRM mix,
equal-domain-combined-std Huber, \texttt{rollout\_sel} checkpoint selection, AdamW
$3\mathrm{e}{-4}\!\to\!3\mathrm{e}{-5}$, $80\times2000$ steps, single seed). The two
tests probe different parts of the design: the first removes the terrain-regime
context supplied to the model, the second changes the reduced-state definition
itself---both the recurrent state and the prediction target.

\begin{itemize}
  \item \textbf{Context ablation} (\texttt{no\_onehot}): drops the $2$-D terrain key,
  input $20\text{-D}\!\to\!18\text{-D}$, readout unchanged at $15$; the model must
  infer rigid-vs-CRM from the state history alone.
  \item \textbf{State-abstraction ablation} (\texttt{no\_tireforce\_omega}): drops the
  four tire normal forces $F_z^{(i)}$ and four wheel speeds $\omega^{(i)}$ from both
  the recurrent state and the prediction target, input $20\text{-D}\!\to\!12\text{-D}$,
  readout $15\!\to\!7$ (body kinematics only:
  $v_x,v_y,\phi,\theta,\omega_x,\omega_y,\omega_z$). This is an ablation of the
  reduced-state definition, not merely of the model's inputs.
\end{itemize}

\begin{table}[t]
  \centering
  \footnotesize
  \caption{Reduced-state and context ablation at the $8$-layer backbone, metrics at
  each run's own best-validation checkpoint. One-step loss is the channel-reweighted Huber
  validation loss; 10\,s open-loop err/dist is the open-loop rollout position error
  normalized by distance traveled, $12$ episodes/domain.
  \textsuperscript{\dag}\,computed over $7$ readout channels instead of $15$ and not
  comparable to the other two rows (see text); the open-loop column is comparable
  across all three, since it is integrated from $v_x,v_y,\omega_z$, which every
  variant retains.}
  \label{tab:feature-ablation}
  \setlength{\tabcolsep}{4pt}
  \begin{tabularx}{\columnwidth}{@{}Y r r r r@{}}
    \toprule
    Model & \multicolumn{2}{c}{One-step loss} & \multicolumn{2}{c}{10\,s open-loop err/dist} \\
    \cmidrule(lr){2-3}\cmidrule(lr){4-5}
    & Flat & CRM & Flat & CRM \\
    \midrule
    \textbf{Baseline (20-D)}   & \textbf{0.00248} & \textbf{0.09378} & \textbf{0.0373} & \textbf{0.0538} \\
    No one-hot (18-D)          & 0.00236          & 0.09395          & 0.0826          & 0.0486 \\
    No terramechanics (12-D)   & 0.00341          & 0.06198\textsuperscript{\dag} & 0.0341 & 0.0800 \\
    \bottomrule
  \end{tabularx}
\end{table}

Removing the one-hot barely changes one-step loss (flat $0.00248\!\to\!0.00236$,
CRM $0.09378\!\to\!0.09395$) but more than doubles open-loop flat error
($0.0373\!\to\!0.0826$) while leaving CRM unchanged. The terrain key is thus largely
invisible to one-step loss but important over long rollouts: without the explicit
regime label, the shared model does not reliably disambiguate the rigid and CRM
transition dynamics from state history alone, producing substantially larger
accumulated error on the rigid domain. Removing the
terramechanics block is the mirror image: flat open-loop error is unaffected
($0.0373\!\to\!0.0341$) but CRM rises $\sim$$49\%$ ($0.0538\!\to\!0.0800$),
consistent with Eq.~\eqref{eq:slip}, since wheel speed and body velocity are
nearly redundant on rigid terrain but decouple under CRM slip. That variant's
one-step CRM loss ($0.062$) is computed over $7$ channels instead of $15$ and is
not comparable to the other rows, so we rank all three on the shared open-loop
metric instead. Both ablations are offline NRD-rollout comparisons excluded from
the architecture ranking of Table~\ref{tab:ofat-sweep}, and the $7$-D checkpoint is
not compatible with the $15$-D state consumed by the RL environment
(Sec.~\ref{subsec:hmmwv-rl}), so we do not report a closed-loop transfer
comparison for it.
\section{Forward-Kinematics Safety Shield for Arm Reaching}
\label{app:arm-safety}

The arm NRD model is trained only on free-space motion---transitions
in contact with the ground, the vehicle, or the arm itself are excluded from its
dataset (Sec.~\ref{subsec:tracked-abstractions})---so it has no notion of contact
and is unreliable for any configuration that would collide. During policy
optimization, where undirected exploration would otherwise drive the arm into such
configurations, we keep every commanded configuration inside the free-space
envelope with a geometric safety shield, and we expose the arm's proximity to
collision to the policy as the clearance channel $\hat\kappa$ of the
observation in Eq.~\eqref{eq:arm-obs}. Both are computed by a batched forward-kinematics
(FK) model that runs on the GPU alongside the reduced dynamics; neither invokes
Chrono. This appendix details that FK clearance; the per-step rule by which the
shield accepts or rejects a proposed command increment using it is given in
Sec.~\ref{subsec:tracked-policies}.

\subsection{Forward Kinematics and Collision Geometry}

\begin{figure}[!ht]
  \centering
  \includegraphics[width=0.55\textwidth]{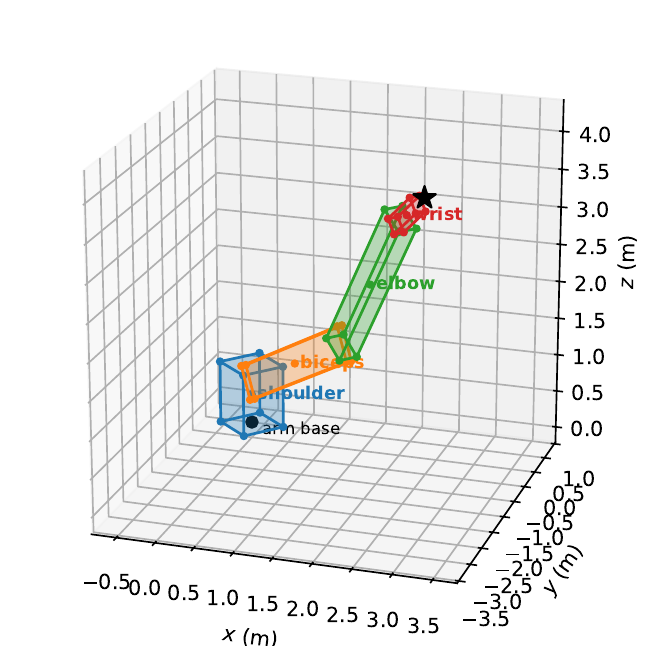}
  \caption{Forward kinematics placing the arm's collision geometry, for an example
  configuration $q=[0.70,0.50,-0.50,0.30]$\,rad. Each of the four arm-chain links
  (shoulder, biceps, elbow, wrist) carries an axis-aligned box that FK maps into
  the world as an \emph{oriented} box; the dots are the eight box corners plus
  center---the sample points $\{x_p(q)\}$ on which the signed-distance
  test in Eq.~\eqref{eq:box-sdf} is evaluated. The star is the end-effector
  $p^{\mathrm{ee}}$, and coordinates are in the arm-base frame. The gripper links
  (end-effector body and two fingers) are omitted for clarity.}
  \label{fig:arm-fk-boxes}
\end{figure}

The $4$-DOF arm is a CAD import with non-DH joint frames, so its kinematics are
extracted once from the Chrono scene and reproduced as a product-of-exponentials
FK evaluated batched over all parallel rollouts. Unlike the end-effector-only
kinematics that the reward needs, the shield needs \emph{every} link pose: the FK
returns the world pose of all $L=7$ collision links (shoulder, biceps, elbow,
wrist, end-effector, and the two fingers). Each link is bounded by an axis-aligned
box, sampled at its eight corners and center; the mounted vehicle is represented
by a single conservative axis-aligned box (padded by $0.2$\,m), and the ground is
the plane $z=z_0$. Figure~\ref{fig:arm-fk-boxes} shows the four arm-chain link
boxes placed by this FK for an example configuration.

\subsection{Signed Clearance}

For a configuration $q$, FK places the link sample points $\{x_p(q)\}$ in the
world. The signed distance from a point $x$ to an axis-aligned box with center $c$
and half-extents $h$ (expressed in the box frame) is
\begin{equation}
  \label{eq:box-sdf}
  d(x;c,h) = \big\lVert\,(\lvert x-c\rvert - h)_{+}\,\big\rVert_2
    + \min\!\big(\textstyle\max_i(\lvert x_i-c_i\rvert - h_i),\ 0\big),
\end{equation}
where $(\cdot)_{+}=\max(\cdot,0)$ elementwise: the first term is the Euclidean
distance to the surface for a point outside the box, and the second is the
(negative) penetration depth for a point inside. Because the two branches meet at
zero on the box surface, $d$ passes continuously from positive (clear) through
zero (touching) to negative (overlapping)---a graded collision signal rather than
a binary flag. For each box $(c,h)$ this $d$ is evaluated against every sample
point that does \emph{not} belong to that box, batched independently over each arm
and each parallel environment. The raw geometric clearance is the minimum of this
per-point distance over all $9L$ sample points and the three collision
constraints,
\begin{equation}
  \label{eq:arm-clearance}
  \kappa(q) = \min\big(\kappa_{\mathrm{ground}},\ \kappa_{\mathrm{veh}},\
  \kappa_{\mathrm{self}}\big),
\end{equation}
\begin{equation}
  \label{eq:arm-clearance-terms}
  \begin{aligned}
    \kappa_{\mathrm{ground}} &= \min_{p}\ \big(z_p(q) - z_0\big), \\[2pt]
    \kappa_{\mathrm{veh}}    &= \min_{p\in\mathcal{D}}\
       d\big(x_p(q);\, c_{\mathrm{veh}}, h_{\mathrm{veh}}\big), \\[2pt]
    \kappa_{\mathrm{self}}   &= \min_{(i,j)\in\mathcal{N}}\ \min_{p\in i}\
       d\big(x^{\,i\to j}_p(q);\, c_j, h_j\big),
  \end{aligned}
\end{equation}
where $x_p(q)$ are the FK-placed sample points and $z_p$ their heights;
$\mathcal{D}$ is the set of points on the distal links (elbow, wrist,
end-effector, and fingers) that can reach the vehicle box
$(c_{\mathrm{veh}},h_{\mathrm{veh}})$; $\mathcal{N}$ is the set of ordered
non-adjacent link pairs---two links joined by a joint overlap at that joint by
construction and would always register contact, so these permanently touching
(adjacent) pairs are excluded, leaving only pairs that should never meet, ordered
so each is tested in both directions---and $x^{\,i\to j}_p$ denotes link $i$'s
point $x_p$ re-expressed in link $j$'s box-aligned frame, so that the box distance
$d$ of Eq.~\eqref{eq:box-sdf} applies. Ground clearance is the one term that reduces to a
point-to-plane height rather than a box distance. Because each box is sampled only at
its corners and center, a positive $\kappa(q)$ provides a conservative
collision-screening signal rather than an exact collision certificate---edge- or
face-first overlaps could in principle slip between sample points---while $\kappa(q)<0$
flags predicted penetration; the zero-contact Chrono evaluation of
Sec.~\ref{subsec:tracked-transfer} is the empirical validation that the screen holds
in deployment.

\subsection{Clearance Observation}

The policy observes the clipped, normalized clearance
\begin{equation}
  \label{eq:kappa-hat}
  \hat\kappa = \operatorname{clip}\!\big(\kappa(q),\,-\kappa_{\mathrm{c}},\,
  \kappa_{\mathrm{c}}\big)\,/\,\kappa_{\mathrm{s}},
\end{equation}
with clip level $\kappa_{\mathrm{c}}=1.0$\,m and scale $\kappa_{\mathrm{s}}=0.5$\,m.
This is effectively an FK-derived proximity sensor: it tells the policy how close
the current configuration is to contact and saturates once the arm is comfortably
far ($>1$\,m) from every obstacle---so the policy can learn collision-aware
behavior without the NRD model ever having to represent obstacles.

Concretely, $\hat\kappa$ enters through a forward-kinematics side-channel wrapped
around the frozen NRD model, evaluated once per control step. Given the
current reduced state, the policy proposes an increment $\Delta q^{\mathrm{cmd}}$;
after interpolation, collision screening, and joint-limit clipping, the accepted
increment updates the absolute setpoint $q^{\mathrm{cmd}}$, and the dynamics model
receives this absolute setpoint as its action and predicts the next reduced state.
Its four joint angles $q$ are read out, FK
re-derives the end-effector position and the link boxes from them, and $\kappa(q)$
---normalized to $\hat\kappa$ by Eq.~\eqref{eq:kappa-hat}---is written into the
observation the policy receives on the following step. The NRD model itself
therefore never represents the ground, the vehicle, or self-contact: obstacle
geometry is injected only through $\hat\kappa$, recomputed by FK on the model's own
predicted joints at every step.
\section{Chrono Policy Evaluation for Tracked Vehicle and Robot Arm}
\label{app:stress}

The closed-loop transfer results in Sec.~\ref{subsec:tracked-transfer} come from
seeded $100$-goal Chrono stress batteries---one per policy---that replace the small
spot checks used in early development. Goals are sampled with a fixed seed
($12345$) from each policy's own trained region, and \emph{each goal is run in its
own freshly built vehicle-plus-arm Chrono process}, so no state leaks between goals
and the arm's per-goal scene rebuild cannot accumulate; every per-goal trajectory
is saved, so success at a tighter tolerance can be recomputed offline without
rerunning Chrono. The headline metrics for both batteries are summarized in
Table~\ref{tab:tracked-arm-transfer}; this appendix shows the underlying
trajectories.

\begin{figure}[!ht]
  \centering
  \includegraphics[width=0.95\textwidth]{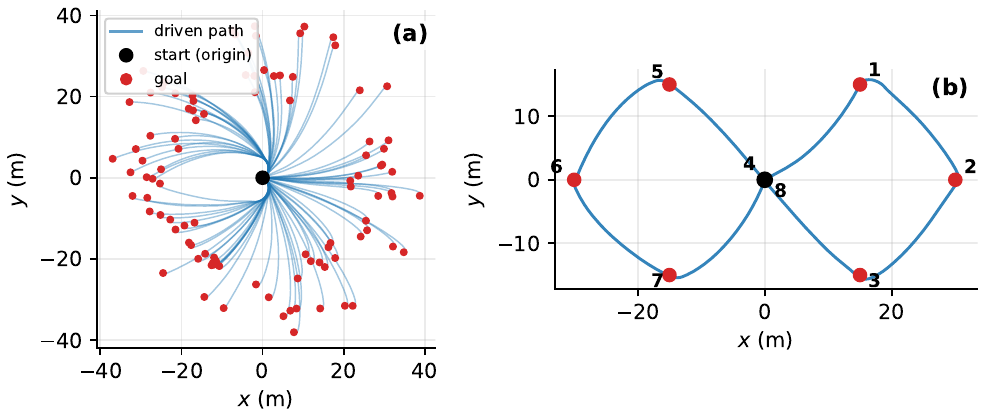}
  \caption{Tracked-vehicle goal reaching in Chrono under the frozen NRD model policy.
  (a)~All $100$ driven paths from the seeded stress battery, each from the common
  origin (black) to its goal (red). (b)~The same policy fed eight consecutive goals
  along a figure-$8$ route, each goal numbered by its order in the sequence (goals
  $4$ and $8$ return through the start).}
  \label{fig:tracked-stress}
\end{figure}

\paragraph{Tracked vehicle.} Figure~\ref{fig:tracked-stress}(a) overlays all $100$
driven paths from the stress battery. The policy reaches every goal across the full
$360^\circ$ distribution, including goals behind the vehicle, which it services with
forward-loop U-turns rather than in-place spins; the paths are near-direct (median
path efficiency $0.959$). It stops as soon as it enters the $0.75$\,m tolerance, so
it hugs that radius---median closest approach $0.691$\,m, with $40/100$ landing in
the $0.70$--$0.75$\,m band---rather than driving all the way onto the goal. Because
each goal in the battery starts from the common origin, we additionally chain the
\emph{same} policy across a fixed sequence of eight goals, resetting only the target
between legs (Fig.~\ref{fig:tracked-stress}(b)). Fed the eight vertices of a
figure-$8$ route spanning $\pm30$\,m, the point-to-point controller reaches all
eight in order ($8/8$, per-leg closest approach $0.46$--$0.69$\,m, $95.5$\,s total)
without any re-planning or per-goal re-tuning---so the single learned goal-reaching
skill composes directly into a waypoint-following controller.

\begin{figure*}[!t]
  \centering
  \includegraphics[width=\textwidth]{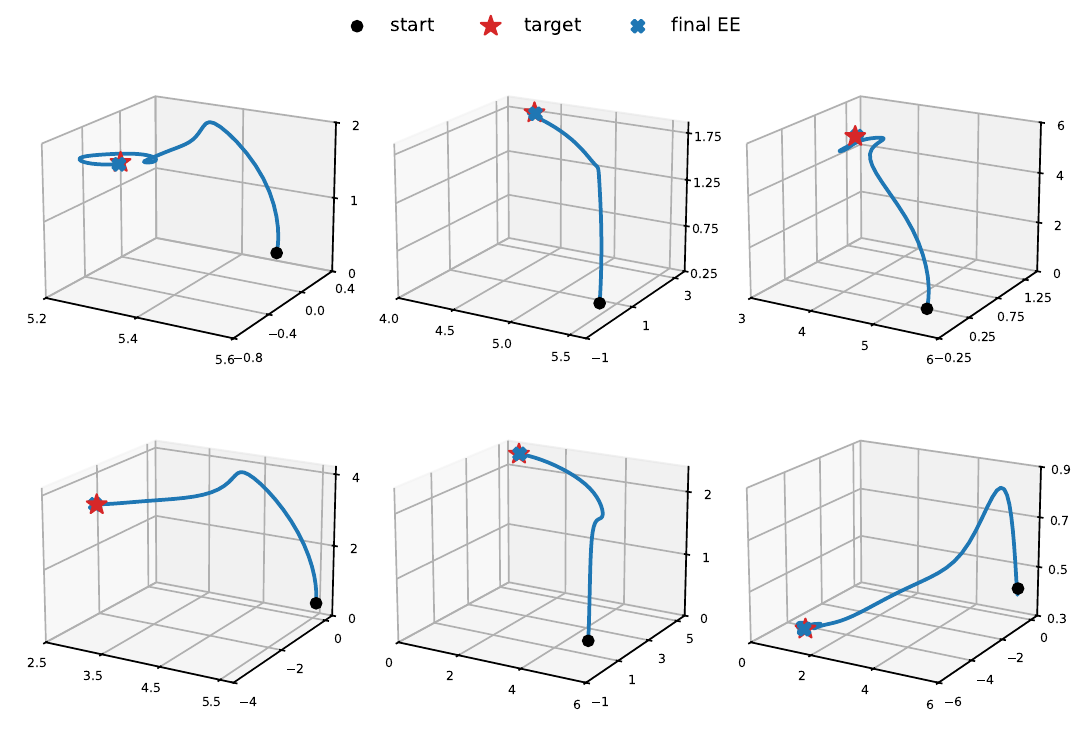}
  \caption{Six representative successful arm reaches from the seeded $100$-goal
  Chrono stress battery, chosen for diverse movement direction and reach length
  (the arm always starts from the same home pose). Each panel is the $3$-D
  end-effector path from the start (circle) to the target (star), with the final
  end-effector position ($\times$); coordinates are in the arm-base frame (drawn
  z-up).}
  \label{fig:arm-stress}
\end{figure*}

\paragraph{Arm.} Figure~\ref{fig:arm-stress} shows six representative successful
reaches chosen for diverse movement---the arm always starts from the same home
pose, so the panels differ in reach direction and length (short near reaches, tall
vertical extensions, and long lateral and near-horizontal sweeps), each converging
the end-effector onto its target within a few centimeters. Across all $100$ goals
the policy reaches $97$ at the
$0.05$\,m tolerance; the three failures are all timeouts at deep lower-workspace
goals the training data under-samples (closest approach $6.4$--$10.9$\,cm), and no
rollout records a contact or joint-limit violation---confirming that the
forward-kinematics safety shield (Appendix~\ref{app:arm-safety}) carries from
training into high-fidelity Chrono.

\bibliographystyle{elsarticle-num-names}
\bibliography{BibFiles/refsGraphics,BibFiles/refsSensors,BibFiles/refsAutonomousVehicles,BibFiles/refsChronoSpecific,BibFiles/refsDEM,BibFiles/refsFSI,BibFiles/refsMBS,BibFiles/refsRobotics,BibFiles/refsSBELspecific,BibFiles/refsTerramech,BibFiles/refsCompSci,BibFiles/refsNumericalIntegr,BibFiles/refsMLPhysics,BibFiles/refsSurfaceTension,BibFiles/refsStatsML,BibFiles/refsOddsEnds,BibFiles/refsML-AI}

\def\cprime{$'$}
\begin{thebibliography}{38}
\expandafter\ifx\csname natexlab\endcsname\relax\def\natexlab#1{#1}\fi
\providecommand{\url}[1]{\texttt{#1}}
\providecommand{\href}[2]{#2}
\providecommand{\path}[1]{#1}
\providecommand{\DOIprefix}{doi:}
\providecommand{\ArXivprefix}{arXiv:}
\providecommand{\URLprefix}{URL: }
\providecommand{\Pubmedprefix}{pmid:}
\providecommand{\doi}[1]{\href{http://dx.doi.org/#1}{\path{#1}}}
\providecommand{\Pubmed}[1]{\href{pmid:#1}{\path{#1}}}
\providecommand{\bibinfo}[2]{#2}
\ifx\xfnm\relax \def\xfnm[#1]{\unskip,\space#1}\fi
\bibitem[{Mazhar et~al.(2013)Mazhar, Heyn, Pazouki, Melanz, Seidl, Bartholomew,
  Tasora, and Negrut}]{chronoOverview2013}
\bibinfo{author}{H.~Mazhar}, \bibinfo{author}{T.~Heyn},
  \bibinfo{author}{A.~Pazouki}, \bibinfo{author}{D.~Melanz},
  \bibinfo{author}{A.~Seidl}, \bibinfo{author}{A.~Bartholomew},
  \bibinfo{author}{A.~Tasora}, \bibinfo{author}{D.~Negrut},
\newblock \bibinfo{title}{{Chrono}: a parallel multi-physics library for
  rigid-body, flexible-body, and fluid dynamics},
\newblock \bibinfo{journal}{Mechanical Sciences} \bibinfo{volume}{4}
  (\bibinfo{year}{2013}) \bibinfo{pages}{49--64}.
  \DOIprefix\doi{10.5194/ms-4-49-2013}.
\bibitem[{Tasora et~al.(2016)Tasora, Serban, Mazhar, Pazouki, Melanz,
  Fleischmann, Taylor, Sugiyama, and Negrut}]{chronoOverview2016}
\bibinfo{author}{A.~Tasora}, \bibinfo{author}{R.~Serban},
  \bibinfo{author}{H.~Mazhar}, \bibinfo{author}{A.~Pazouki},
  \bibinfo{author}{D.~Melanz}, \bibinfo{author}{J.~Fleischmann},
  \bibinfo{author}{M.~Taylor}, \bibinfo{author}{H.~Sugiyama},
  \bibinfo{author}{D.~Negrut},
\newblock \bibinfo{title}{Chrono: An open source multi-physics dynamics
  engine},
\newblock in: \bibinfo{editor}{T.~Kozubek}, \bibinfo{editor}{R.~Blaheta},
  \bibinfo{editor}{J.~{\v{S}}{\'i}stek},
  \bibinfo{editor}{M.~Rozlo{\v{z}}n{\'i}k},
  \bibinfo{editor}{M.~{\v{C}}erm{\'a}k} (Eds.), \bibinfo{booktitle}{High
  Performance Computing in Science and Engineering},
  \bibinfo{publisher}{Springer International Publishing},
  \bibinfo{address}{Cham}, \bibinfo{year}{2016}, pp. \bibinfo{pages}{19--49}.
  \DOIprefix\doi{10.1007/978-3-319-40361-8_2}.
\bibitem[{Serban et~al.(2019)Serban, Taylor, Negrut, and
  Tasora}]{chronoVehicle2019}
\bibinfo{author}{R.~Serban}, \bibinfo{author}{M.~Taylor},
  \bibinfo{author}{D.~Negrut}, \bibinfo{author}{A.~Tasora},
\newblock \bibinfo{title}{{Chrono::Vehicle} template-based ground vehicle
  modeling and simulation},
\newblock \bibinfo{journal}{International Journal of Vehicle Performance}
  \bibinfo{volume}{5} (\bibinfo{year}{2019}) \bibinfo{pages}{18--39}.
  \DOIprefix\doi{10.1504/IJVP.2019.097096}.
\bibitem[{Wang et~al.(2026)Wang, Wang, and Negrut}]{justin-FNODE-2026}
\bibinfo{author}{H.~Wang}, \bibinfo{author}{J.~Wang},
  \bibinfo{author}{D.~Negrut},
\newblock \bibinfo{title}{{FNODE}: {F}low-matching for data-driven simulation
  of constrained multibody systems},
\newblock \bibinfo{journal}{Computer Methods in Applied Mechanics and
  Engineering} \bibinfo{volume}{455} (\bibinfo{year}{2026})
  \bibinfo{pages}{118912}. \URLprefix
  \url{https://www.sciencedirect.com/science/article/pii/S0045782526001854}.
  \DOIprefix\doi{https://doi.org/10.1016/j.cma.2026.118912}.
\bibitem[{Vaswani et~al.(2017)Vaswani, Shazeer, Parmar, Uszkoreit, Jones,
  Gomez, Kaiser, and Polosukhin}]{vaswani2017attention}
\bibinfo{author}{A.~Vaswani}, \bibinfo{author}{N.~Shazeer},
  \bibinfo{author}{N.~Parmar}, \bibinfo{author}{J.~Uszkoreit},
  \bibinfo{author}{L.~Jones}, \bibinfo{author}{A.~N. Gomez},
  \bibinfo{author}{{\L}.~Kaiser}, \bibinfo{author}{I.~Polosukhin},
\newblock \bibinfo{title}{Attention is all you need},
\newblock \bibinfo{journal}{Advances in neural information processing systems}
  \bibinfo{volume}{30} (\bibinfo{year}{2017}).
\bibitem[{Schulman et~al.(2017)Schulman, Wolski, Dhariwal, Radford, and
  Klimov}]{schulman2017proximal}
\bibinfo{author}{J.~Schulman}, \bibinfo{author}{F.~Wolski},
  \bibinfo{author}{P.~Dhariwal}, \bibinfo{author}{A.~Radford},
  \bibinfo{author}{O.~Klimov},
\newblock \bibinfo{title}{Proximal policy optimization algorithms},
\newblock \bibinfo{journal}{arXiv preprint arXiv:1707.06347}
  (\bibinfo{year}{2017}).
\bibitem[{Hu et~al.(2021)Hu, Rakhsha, Yang, Kamrin, and
  Negrut}]{weiGranularSPH2021}
\bibinfo{author}{W.~Hu}, \bibinfo{author}{M.~Rakhsha},
  \bibinfo{author}{L.~Yang}, \bibinfo{author}{K.~Kamrin},
  \bibinfo{author}{D.~Negrut},
\newblock \bibinfo{title}{Modeling granular material dynamics and its two-way
  coupling with moving solid bodies using a continuum representation and the
  {SPH} method},
\newblock \bibinfo{journal}{Computer Methods in Applied Mechanics and
  Engineering} \bibinfo{volume}{385} (\bibinfo{year}{2021})
  \bibinfo{pages}{114022}. \DOIprefix\doi{10.1016/j.cma.2021.114022}.
\bibitem[{Unjhawala et~al.(2026)Unjhawala, Bakke, Zhang, Taylor, Arivoli,
  Serban, and Negrut}]{Huzaifa2026CRM}
\bibinfo{author}{H.~M. Unjhawala}, \bibinfo{author}{L.~Bakke},
  \bibinfo{author}{H.~Zhang}, \bibinfo{author}{M.~Taylor},
  \bibinfo{author}{G.~Arivoli}, \bibinfo{author}{R.~Serban},
  \bibinfo{author}{D.~Negrut},
\newblock \bibinfo{title}{A physics-based continuum model for versatile,
  scalable, and fast terramechanics simulation},
\newblock \bibinfo{journal}{Journal of Terramechanics} \bibinfo{volume}{124}
  (\bibinfo{year}{2026}) \bibinfo{pages}{101150}. \URLprefix
  \url{https://www.sciencedirect.com/science/article/pii/S0022489826000339}.
  \DOIprefix\doi{https://doi.org/10.1016/j.jterra.2026.101150}.
\bibitem[{Chua et~al.(2018)Chua, Calandra, McAllister, and
  Levine}]{chua2018pets}
\bibinfo{author}{K.~Chua}, \bibinfo{author}{R.~Calandra},
  \bibinfo{author}{R.~McAllister}, \bibinfo{author}{S.~Levine},
\newblock \bibinfo{title}{Deep reinforcement learning in a handful of trials
  using probabilistic dynamics models},
\newblock in: \bibinfo{booktitle}{Advances in Neural Information Processing
  Systems (NeurIPS)}, volume~\bibinfo{volume}{31}, \bibinfo{year}{2018}, pp.
  \bibinfo{pages}{4754--4765}. \href{http://arxiv.org/abs/1805.12114}{{\tt
  arXiv:1805.12114}}.
\bibitem[{Janner et~al.(2019)Janner, Fu, Zhang, and Levine}]{janner2019mbpo}
\bibinfo{author}{M.~Janner}, \bibinfo{author}{J.~Fu},
  \bibinfo{author}{M.~Zhang}, \bibinfo{author}{S.~Levine},
\newblock \bibinfo{title}{When to trust your model: Model-based policy
  optimization},
\newblock in: \bibinfo{booktitle}{Advances in Neural Information Processing
  Systems (NeurIPS)}, volume~\bibinfo{volume}{32}, \bibinfo{year}{2019}, pp.
  \bibinfo{pages}{12519--12530}. \href{http://arxiv.org/abs/1906.08253}{{\tt
  arXiv:1906.08253}}.
\bibitem[{Ha and Schmidhuber(2018)}]{ha2018worldmodels}
\bibinfo{author}{D.~Ha}, \bibinfo{author}{J.~Schmidhuber},
  \bibinfo{title}{World models}, \bibinfo{year}{2018}. \URLprefix
  \url{https://arxiv.org/abs/1803.10122}.
  \DOIprefix\doi{10.48550/arXiv.1803.10122}.
  \href{http://arxiv.org/abs/1803.10122}{{\tt arXiv:1803.10122}}.
\bibitem[{Hafner et~al.(2019)Hafner, Lillicrap, Fischer, Villegas, Ha, Lee, and
  Davidson}]{hafner2019planet}
\bibinfo{author}{D.~Hafner}, \bibinfo{author}{T.~Lillicrap},
  \bibinfo{author}{I.~Fischer}, \bibinfo{author}{R.~Villegas},
  \bibinfo{author}{D.~Ha}, \bibinfo{author}{H.~Lee},
  \bibinfo{author}{J.~Davidson},
\newblock \bibinfo{title}{Learning latent dynamics for planning from pixels},
\newblock in: \bibinfo{booktitle}{Proceedings of the 36th International
  Conference on Machine Learning (ICML)}, \bibinfo{year}{2019}, pp.
  \bibinfo{pages}{2555--2565}. \href{http://arxiv.org/abs/1811.04551}{{\tt
  arXiv:1811.04551}}.
\bibitem[{Hafner et~al.(2020)Hafner, Lillicrap, Ba, and
  Norouzi}]{hafner2020dreamer}
\bibinfo{author}{D.~Hafner}, \bibinfo{author}{T.~Lillicrap},
  \bibinfo{author}{J.~Ba}, \bibinfo{author}{M.~Norouzi},
\newblock \bibinfo{title}{Dream to control: Learning behaviors by latent
  imagination},
\newblock in: \bibinfo{booktitle}{International Conference on Learning
  Representations (ICLR)}, \bibinfo{year}{2020}.
\bibitem[{Levy et~al.(2026)Levy, Westenbroek, Huang, Palafox, Yin, Omidshafiei,
  Kim, Gupta, and Fridovich-Keil}]{levy2026simdist}
\bibinfo{author}{J.~Levy}, \bibinfo{author}{T.~Westenbroek},
  \bibinfo{author}{K.~Huang}, \bibinfo{author}{F.~Palafox},
  \bibinfo{author}{P.~Yin}, \bibinfo{author}{S.~Omidshafiei},
  \bibinfo{author}{D.-K. Kim}, \bibinfo{author}{A.~Gupta},
  \bibinfo{author}{D.~Fridovich-Keil},
\newblock \bibinfo{title}{Simulation distillation: Pretraining world models in
  simulation for rapid real-world adaptation},
\newblock \bibinfo{journal}{arXiv preprint arXiv:2603.15759}
  (\bibinfo{year}{2026}). \bibinfo{note}{To appear at Robotics: Science and
  Systems (RSS) 2026}.
\bibitem[{Xu et~al.(2025)Xu, Heiden, Akinola, Fox, Macklin, and
  Narang}]{xu2025nerd}
\bibinfo{author}{J.~Xu}, \bibinfo{author}{E.~Heiden},
  \bibinfo{author}{I.~Akinola}, \bibinfo{author}{D.~Fox},
  \bibinfo{author}{M.~Macklin}, \bibinfo{author}{Y.~Narang},
\newblock \bibinfo{title}{Neural robot dynamics},
\newblock \bibinfo{journal}{arXiv preprint arXiv:2508.15755}
  (\bibinfo{year}{2025}). \bibinfo{note}{Published at Conference on Robot
  Learning (CoRL) 2025}.
\bibitem[{Moore et~al.(2026)Moore, Lee, and Chen}]{moore2026snsmpc}
\bibinfo{author}{S.~A. Moore}, \bibinfo{author}{E.~Lee},
  \bibinfo{author}{B.~Chen},
\newblock \bibinfo{title}{Learning legged mpc with smooth neural surrogates},
\newblock \bibinfo{journal}{arXiv preprint arXiv:2601.12169}
  (\bibinfo{year}{2026}).
\bibitem[{Altawaitan and Atanasov(2026)}]{altawaitan2026adapting}
\bibinfo{author}{A.~Altawaitan}, \bibinfo{author}{N.~Atanasov},
\newblock \bibinfo{title}{Adapting neural robot dynamics on the fly for
  predictive control},
\newblock \bibinfo{journal}{arXiv preprint arXiv:2604.04039}
  (\bibinfo{year}{2026}).
\bibitem[{Jing et~al.(2026)Jing, Bandi, Ye, Duan, Abbeel, Wang, and
  Yi}]{jing2026contactaware}
\bibinfo{author}{C.~Jing}, \bibinfo{author}{J.~K. Bandi},
  \bibinfo{author}{J.~Ye}, \bibinfo{author}{Y.~Duan},
  \bibinfo{author}{P.~Abbeel}, \bibinfo{author}{X.~Wang},
  \bibinfo{author}{S.~Yi},
\newblock \bibinfo{title}{Contact-aware neural dynamics},
\newblock \bibinfo{journal}{arXiv preprint arXiv:2601.12796}
  (\bibinfo{year}{2026}).
\bibitem[{Benner et~al.(2015)Benner, Gugercin, and Willcox}]{benner2015survey}
\bibinfo{author}{P.~Benner}, \bibinfo{author}{S.~Gugercin},
  \bibinfo{author}{K.~Willcox},
\newblock \bibinfo{title}{A survey of projection-based model reduction methods
  for parametric dynamical systems},
\newblock \bibinfo{journal}{SIAM Review} \bibinfo{volume}{57}
  (\bibinfo{year}{2015}) \bibinfo{pages}{483--531}.
  \DOIprefix\doi{10.1137/130932715}.
\bibitem[{Xiang et~al.(2026)Xiang, Modi, Dagli, Trusty, Daviet, Chen, Sharp,
  and Levin}]{xiang2026freeform}
\bibinfo{author}{D.~Xiang}, \bibinfo{author}{V.~Modi},
  \bibinfo{author}{R.~Dagli}, \bibinfo{author}{T.~Trusty},
  \bibinfo{author}{G.~Daviet}, \bibinfo{author}{A.~H. Chen},
  \bibinfo{author}{N.~Sharp}, \bibinfo{author}{D.~I.~W. Levin},
\newblock \bibinfo{title}{Freeform: Reduced-order deformable simulation from
  particle-based skinning eigenmodes},
\newblock \bibinfo{journal}{arXiv preprint arXiv:2605.29318}
  (\bibinfo{year}{2026}). \bibinfo{note}{To appear at CVPR 2026}.
\bibitem[{Romero et~al.(2021)Romero, Casas, P{\'e}rez, and
  Otaduy}]{romero2021contactcorrections}
\bibinfo{author}{C.~Romero}, \bibinfo{author}{D.~Casas},
  \bibinfo{author}{J.~P{\'e}rez}, \bibinfo{author}{M.~A. Otaduy},
\newblock \bibinfo{title}{Learning contact corrections for handle-based
  subspace dynamics},
\newblock \bibinfo{journal}{ACM Transactions on Graphics (Proc. SIGGRAPH)}
  \bibinfo{volume}{40} (\bibinfo{year}{2021}).
  \DOIprefix\doi{10.1145/3450626.3459875}.
\bibitem[{Ly et~al.(2025)Ly, Tatsuoka, Nagaraj, Levy, Palafox, Fridovich-Keil,
  and Lu}]{ly2025datadriven}
\bibinfo{author}{N.~Ly}, \bibinfo{author}{C.~Tatsuoka},
  \bibinfo{author}{J.~Nagaraj}, \bibinfo{author}{J.~Levy},
  \bibinfo{author}{F.~Palafox}, \bibinfo{author}{D.~Fridovich-Keil},
  \bibinfo{author}{H.~Lu},
\newblock \bibinfo{title}{Data-driven modeling and correction of vehicle
  dynamics},
\newblock \bibinfo{journal}{arXiv preprint arXiv:2512.00289}
  (\bibinfo{year}{2025}). \bibinfo{note}{Also published in Journal of Machine
  Learning for Modeling and Computing, vol. 7, issue 2, 2026}.
\bibitem[{Otto and Rowley(2019)}]{otto2019lran}
\bibinfo{author}{S.~E. Otto}, \bibinfo{author}{C.~W. Rowley},
\newblock \bibinfo{title}{Linearly recurrent autoencoder networks for learning
  dynamics},
\newblock \bibinfo{journal}{SIAM Journal on Applied Dynamical Systems}
  \bibinfo{volume}{18} (\bibinfo{year}{2019}) \bibinfo{pages}{558--593}.
  \href{http://arxiv.org/abs/1712.01378}{{\tt arXiv:1712.01378}}.
\bibitem[{Tasora et~al.(2019)Tasora, Mangoni, Negrut, Serban, and
  Jayakumar}]{chronoSCM2019}
\bibinfo{author}{A.~Tasora}, \bibinfo{author}{D.~Mangoni},
  \bibinfo{author}{D.~Negrut}, \bibinfo{author}{R.~Serban},
  \bibinfo{author}{P.~Jayakumar},
\newblock \bibinfo{title}{Deformable soil with adaptive level of detail for
  tracked and wheeled vehicles},
\newblock \bibinfo{journal}{International Journal of Vehicle Performance}
  \bibinfo{volume}{5} (\bibinfo{year}{2019}) \bibinfo{pages}{60--76}.
  \DOIprefix\doi{10.1504/IJVP.2019.097098}.
\bibitem[{Todorov et~al.(2012)Todorov, Erez, and Tassa}]{todorovMujoco2012}
\bibinfo{author}{E.~Todorov}, \bibinfo{author}{T.~Erez},
  \bibinfo{author}{Y.~Tassa},
\newblock \bibinfo{title}{{MuJoCo}: A physics engine for model-based control},
\newblock in: \bibinfo{booktitle}{2012 IEEE/RSJ International Conference on
  Intelligent Robots and Systems}, \bibinfo{organization}{IEEE},
  \bibinfo{year}{2012}, pp. \bibinfo{pages}{5026--5033}.
\bibitem[{Zakka et~al.(2025)Zakka, Tabanpour, Liao, Haiderbhai, Holt, Luo,
  Allshire, Frey, Sreenath, Kahrs, Sferrazza, Tassa, and
  Abbeel}]{mujoco_playground_2025}
\bibinfo{author}{K.~Zakka}, \bibinfo{author}{B.~Tabanpour},
  \bibinfo{author}{Q.~Liao}, \bibinfo{author}{M.~Haiderbhai},
  \bibinfo{author}{S.~Holt}, \bibinfo{author}{J.~Y. Luo},
  \bibinfo{author}{A.~Allshire}, \bibinfo{author}{E.~Frey},
  \bibinfo{author}{K.~Sreenath}, \bibinfo{author}{L.~A. Kahrs},
  \bibinfo{author}{C.~Sferrazza}, \bibinfo{author}{Y.~Tassa},
  \bibinfo{author}{P.~Abbeel}, \bibinfo{title}{Mujoco playground: An
  open-source framework for gpu-accelerated robot learning and sim-to-real
  transfer}, \bibinfo{year}{2025}. \URLprefix
  \url{https://arxiv.org/abs/2502.08844}.
  \href{http://arxiv.org/abs/2502.08844}{{\tt arXiv:2502.08844}}.
\bibitem[{{NVIDIA}(2025)}]{isaaclab2025}
\bibinfo{author}{{NVIDIA}}, \bibinfo{title}{Isaac lab: Unified framework for
  robot learning built on nvidia isaac sim}, \bibinfo{year}{2025}. \URLprefix
  \url{https://github.com/isaac-sim/IsaacLab}, \bibinfo{note}{accessed
  2026-04-29}.
\bibitem[{Freeman et~al.(2021)Freeman, Frey, Raichuk, Girgin, Mordatch, and
  Bachem}]{brax-freeman2021}
\bibinfo{author}{C.~D. Freeman}, \bibinfo{author}{E.~Frey},
  \bibinfo{author}{A.~Raichuk}, \bibinfo{author}{S.~Girgin},
  \bibinfo{author}{I.~Mordatch}, \bibinfo{author}{O.~Bachem},
\newblock \bibinfo{title}{Brax: A differentiable physics engine for large scale
  rigid body simulation},
\newblock in: \bibinfo{booktitle}{Proceedings of the Neural Information
  Processing Systems Track on Datasets and Benchmarks}, \bibinfo{year}{2021}.
  \URLprefix \url{https://github.com/google/brax}.
  \href{http://arxiv.org/abs/2106.13281}{{\tt arXiv:2106.13281}}.
\bibitem[{Authors(2024)}]{genesisSimulator}
\bibinfo{author}{G.~Authors}, \bibinfo{title}{Genesis: A universal and
  generative physics engine for robotics and beyond},
  \bibinfo{howpublished}{\url{https://github.com/Genesis-Embodied-AI/Genesis}},
  \bibinfo{year}{2024}. \bibinfo{note}{Accessed: [January 5, 2025]}.
\bibitem[{Lee et~al.(2023)Lee, Kim, Mun, and Lee}]{lee2023terrainaware}
\bibinfo{author}{H.~Lee}, \bibinfo{author}{T.~Kim}, \bibinfo{author}{J.~Mun},
  \bibinfo{author}{W.~Lee},
\newblock \bibinfo{title}{Learning terrain-aware kinodynamic model for
  autonomous off-road rally driving with model predictive path integral
  control},
\newblock \bibinfo{journal}{IEEE Robotics and Automation Letters}
  \bibinfo{volume}{8} (\bibinfo{year}{2023}) \bibinfo{pages}{7663--7670}.
  \DOIprefix\doi{10.1109/LRA.2023.3318190}.
  \href{http://arxiv.org/abs/2305.00676}{{\tt arXiv:2305.00676}}.
\bibitem[{Gibson et~al.(2026)Gibson, Vlahov, Spieler, and
  Theodorou}]{gibson2026multistep}
\bibinfo{author}{J.~Gibson}, \bibinfo{author}{B.~Vlahov},
  \bibinfo{author}{P.~Spieler}, \bibinfo{author}{E.~A. Theodorou},
\newblock \bibinfo{title}{Multistep belief space dynamics learning for
  risk-aware control},
\newblock \bibinfo{journal}{arXiv preprint arXiv:2605.12628}
  (\bibinfo{year}{2026}).
\bibitem[{Amine et~al.(2026)Amine, Puri, Le, and
  Mangharam}]{amine2026nonplanar}
\bibinfo{author}{A.~Amine}, \bibinfo{author}{K.~R. Puri},
  \bibinfo{author}{V.-A. Le}, \bibinfo{author}{R.~Mangharam},
\newblock \bibinfo{title}{Nonplanar model predictive control for autonomous
  vehicles with recursive sparse gaussian process dynamics},
\newblock in: \bibinfo{booktitle}{2026 IEEE Intelligent Vehicles Symposium
  (IV)}, \bibinfo{organization}{IEEE}, \bibinfo{year}{2026}.
  \href{http://arxiv.org/abs/2602.16206}{{\tt arXiv:2602.16206}}.
\bibitem[{Baxter et~al.(2026)Baxter, Epureanu, Jayakumar, and
  Ersal}]{baxter2026highspeed}
\bibinfo{author}{J.~R. Baxter}, \bibinfo{author}{B.~I. Epureanu},
  \bibinfo{author}{P.~Jayakumar}, \bibinfo{author}{T.~Ersal},
\newblock \bibinfo{title}{High-speed, all-terrain autonomy: Ensuring safety at
  the limits of mobility},
\newblock \bibinfo{journal}{arXiv preprint arXiv:2603.20525}
  (\bibinfo{year}{2026}).
\bibitem[{Wu et~al.(2026)Wu, Song, Mundheda, Navarro-Serment, Schoenborn, and
  Schneider}]{wu2026tadpo}
\bibinfo{author}{Z.~Wu}, \bibinfo{author}{R.~Song},
  \bibinfo{author}{V.~Mundheda}, \bibinfo{author}{L.~E. Navarro-Serment},
  \bibinfo{author}{C.~Schoenborn}, \bibinfo{author}{J.~Schneider},
\newblock \bibinfo{title}{Tadpo: Reinforcement learning goes off-road},
\newblock \bibinfo{journal}{arXiv preprint arXiv:2603.05995}
  (\bibinfo{year}{2026}). \bibinfo{note}{Accepted at ICRA 2026}.
\bibitem[{Lee et~al.(2020)Lee, Seo, Lee, Lee, and Shin}]{lee2020contextaware}
\bibinfo{author}{K.~Lee}, \bibinfo{author}{Y.~Seo}, \bibinfo{author}{S.~Lee},
  \bibinfo{author}{H.~Lee}, \bibinfo{author}{J.~Shin},
\newblock \bibinfo{title}{Context-aware dynamics model for generalization in
  model-based reinforcement learning},
\newblock in: \bibinfo{booktitle}{Proceedings of the 37th International
  Conference on Machine Learning (ICML)}, volume \bibinfo{volume}{119},
  \bibinfo{year}{2020}, pp. \bibinfo{pages}{5757--5766}.
  \href{http://arxiv.org/abs/2005.06800}{{\tt arXiv:2005.06800}}.
\bibitem[{Zhang and Negrut(2026)}]{NeDMprojectPage}
\bibinfo{author}{H.~Zhang}, \bibinfo{author}{D.~Negrut}, \bibinfo{title}{{NeDM:
  Neural Reduced Dynamics for Complex Robot Control -- Project Page, Code,
  Dataset, and Videos}},
  \bibinfo{howpublished}{\url{https://uwsbel.github.io/NeDM/}},
  \bibinfo{year}{2026}. \bibinfo{note}{Accessed: 2026-08-19}.
\bibitem[{Rill(2015)}]{Rill15}
\bibinfo{author}{G.~Rill},
\newblock \bibinfo{title}{An engineer's guess on tyre parameter made possible
  with {TMeasy}},
\newblock in: \bibinfo{booktitle}{Proceedings of the 4th International Tyre
  Colloquium in. University of Surrey, GB.
  \url{http://epubs.surrey.ac.uk/807823}}, \bibinfo{year}{2015}.
\bibitem[{Modi et~al.(2024)Modi, Sharp, Perel, Sueda, and
  Levin}]{modi2024simplicits}
\bibinfo{author}{V.~Modi}, \bibinfo{author}{N.~Sharp},
  \bibinfo{author}{O.~Perel}, \bibinfo{author}{S.~Sueda},
  \bibinfo{author}{D.~I.~W. Levin},
\newblock \bibinfo{title}{Simplicits: Mesh-free, geometry-agnostic elastic
  simulation},
\newblock \bibinfo{journal}{ACM Transactions on Graphics (Proc. SIGGRAPH)}
  \bibinfo{volume}{43} (\bibinfo{year}{2024}).
  \DOIprefix\doi{10.1145/3658184.3658200}.

\end{thebibliography}

\end{document}